\documentclass[10pt,twocolumn,letterpaper]{article}

\usepackage[pagenumbers]{cvpr} % To force page numbers, e.g. for an arXiv version

\definecolor{cvprblue}{rgb}{0.21,0.49,0.74}
\usepackage[pagebackref,breaklinks,colorlinks,allcolors=cvprblue]{hyperref}
\usepackage{multicol}
\usepackage{multirow}
\usepackage{tabularx}
\usepackage{adjustbox}
\def\paperID{} % *** Enter the Paper ID here
\def\confName{CVPR}
\def\confYear{2025}

\title{RoGs: Large Scale Road Surface Reconstruction with Meshgrid Gaussian}
\author{%
  Zhiheng Feng  \qquad Wenhua Wu  \qquad Tianchen Deng \qquad Hesheng Wang\thanks{Corresponding Author.} \\
  Shanghai Jiao Tong University\\
}

\begin{document}
\maketitle
\begin{abstract}
Road surface reconstruction plays a crucial role in autonomous driving, which can be used for road lane perception and autolabeling. Recently, mesh-based road surface reconstruction algorithms have shown promising reconstruction results. However, these mesh-based methods suffer from slow speed and poor reconstruction quality. To address these limitations, we propose a novel large-scale road surface reconstruction approach with meshgrid Gaussian, named RoGs. Specifically, we model the road surface by placing Gaussian surfels in the vertices of a uniformly distributed square mesh, where each surfel stores color, semantic, and geometric information. This square mesh-based layout covers the entire road with fewer Gaussian surfels and reduces the overlap between Gaussian surfels during training. In addition, because the road surface has no thickness, 2D Gaussian surfel is more consistent with the physical reality of the road surface than 3D Gaussian sphere. Then, unlike previous initialization methods that rely on point clouds, we introduce a vehicle pose-based initialization method to initialize the height and rotation of the Gaussian surfel. Thanks to this meshgrid Gaussian modeling and pose-based initialization, our method achieves significant speedups while improving reconstruction quality. We obtain excellent results in reconstruction of road surfaces in a variety of challenging real-world scenes.

\end{abstract}    
\section{Introduction}
\label{sec:intro}

% \begin{figure*}
%   \centering
%   \includegraphics[width=1.0\linewidth]{fig/kitti1-zip.png}
%   \caption{We propose RoGs, a large scale road surface reconstruction method based on Gaussian Splatting. Inputting the surround view video, semantic segmentation results and location trajectory, RoGs is able to reconstruct RGB maps, semantic maps and elevation maps.}
%   \label{fig:kitti}
% \end{figure*}

\begin{figure*}
  \centering
  \includegraphics[width=0.95\linewidth]{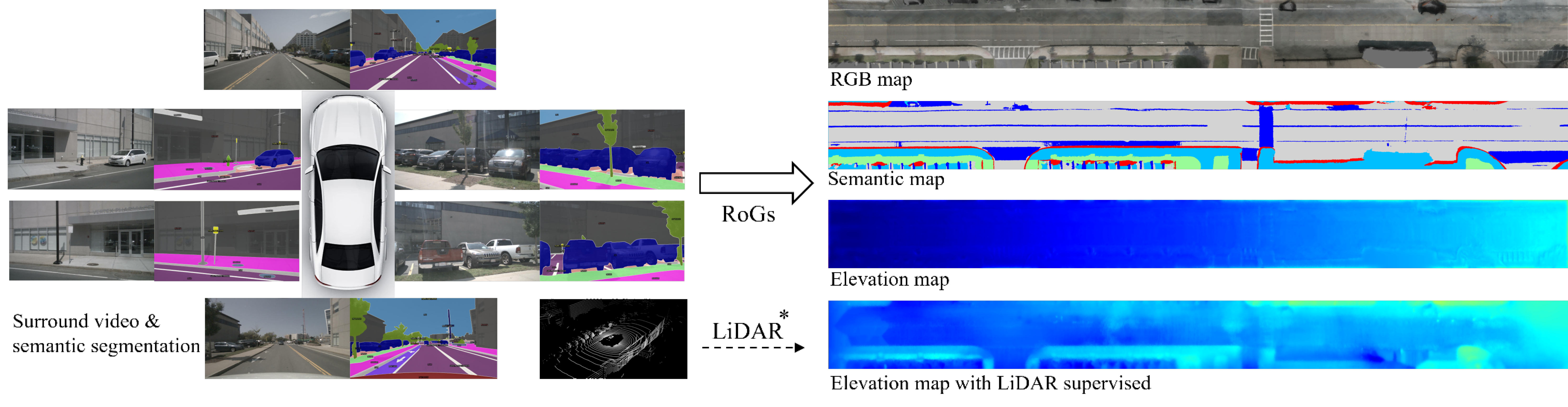}
  \caption{We propose RoGs, a large-scale road surface reconstruction method based on Gaussian Splatting. Inputting the surround view video, semantic segmentation results, and vehicle poses, RoGs is able to reconstruct RGB maps, semantic maps, and elevation maps.}
  \label{fig:pipeline}
\end{figure*}

The development of autonomous driving technology triggers a significant increase in the demand for road reconstruction. Road surface reconstruction tasks involve recovering geometric and texture information of the driving area from video data collected by vehicles. Road surface reconstruction focuses on road regions, lane lines, and road markings. On the one hand, it can be used for high-definition maps and data annotation~\cite{wu2024emie, mei2024rome}. On the other hand, it can provide support for BEV perception~\cite{zhao2024roadbev, zhao2023rsrd}.

Traditional reconstruction methods focus on recovering the overall geometric structure of the scene. Structure-from-Motion (SfM)~\cite{ozyecsil2017survey} and Multi-View Stereo (MVS) methods~\cite{zhou2021dp} can reconstruct sparse or semi-dense point clouds of the scene. However, for road areas with sparse features, there are often holes and noise. Nerf-based methods~\cite{tancik2022block, turki2022mega, barron2022mip, song2023sc, zhu2023sni} can achieve realistic rendering effects but struggle to recover scene geometry from implicit representations, especially for large scenes. The emergence of 3D Gaussian Splatting~\cite{kerbl20233d, chen2024survey} greatly improves rendering speed and has a natural advantage in scene geometry reconstruction due to the approximation of the explicit Gaussian sphere. However, current Gaussian-based reconstruction methods mainly focus on rendering quality and do not fully utilize the explicit features of Gaussian spheres for scene geometry recovery. Recent methods attempt to use explicit meshes or a combination of explicit meshes and implicit encoding for scene representation, achieving decent road surface reconstruction results~\cite{wu2024emie, mei2024rome}. However, the quality and efficiency  are still insufficient for practical applications.

In this context, we propose RoGs, a large-scale road surface reconstruction method with meshgrid Gaussian, which utilizes mesh-distributed Gaussian surfels to achieve efficient and high-quality road surface reconstruction. Specifically, we model the road surface by placing Gaussian surfels in the vertices of a uniformly distributed square mesh, where each surfel stores color, semantic, and geometric information. This square mesh-based layout covers the entire road with fewer Gaussian surfels and reduces the overlap between Gaussian surfels. In addition, because the road surface has no thickness, 2D Gaussian surfel is more consistent with the physical reality of the road surface than 3D Gaussian sphere.  Gaussian surfel is 2D ellipsoidal surface obtained by setting the z-axis scale of a 3D Gaussian sphere to 0. Then, unlike previous initialization methods that use point clouds or random initialization, we propose a vehicle pose-based initialization method. Under the prior condition that the vehicle poses are parallel to the road surface, we initialize the road area range using the vehicle trajectory range and initialize the height and rotation of the Gaussian surfel using the nearest vehicle pose. This initialization method fully leverages the relationship between the vehicle pose and the road surface, which is more conducive to the subsequent optimization of road surface geometric textures. After initialization, the rendering is supervised by camera images and semantic segmentation results. In addition, the height can also be supervised by point clouds. Finally, we obtain 3D road surface reconstruction result with color and semantics. Our contributions are summarized as follows:

\begin{itemize}
\item We propose RoGs, a large-scale road surface reconstruction method with meshgrid Gaussian. The core of RoGs is the mesh-distributed Gaussian surfel representation that conforms to the physical properties of road surfaces.

\item We propose a vehicle pose-based initialization method, which makes the Gaussian surfels closer to the real-world road surface, facilitating optimization.

\item Experiment results on the KITTI and Nuscenes datasets demonstrate that our method achieves efficient and high-quality road surface reconstruction. Importantly, our method improves the speed up by \textbf{53$\times$} and \textbf{27$\times$} when iterating one epoch and two epochs, respectively.
\end{itemize}
\section{Related Works}
\label{sec:relate}

\begin{figure*}[htp]
  \centering
  \includegraphics[width=0.9\linewidth]{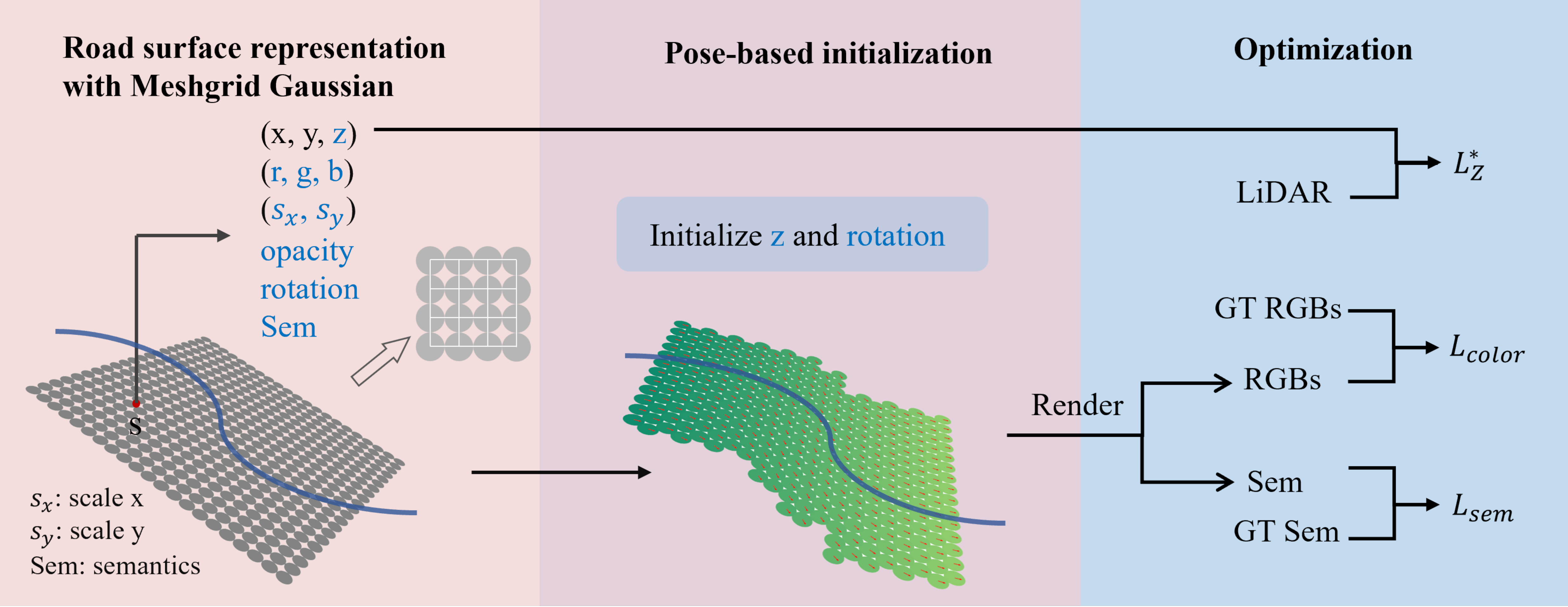}
  \caption{Overview of RoGs. The left side shows the road representation with meshgrid Gaussian. The blue curve indicates the vehicle trajectory. Mesh-distributed Gaussian surfels are used to represent the road surface. Each surfel stores position, scale, rotation, color, opacity, and semantic information. The learnable parameters are indicated in \textcolor[HTML]{0070C0}{blue}. The middle demonstrates the pose-based initialization. For each Gaussian surfel, its elevation (z) and rotation are initialized using the nearest vehicle pose. Finally, the rendering is supervised by camera images and semantic segmentation results. Additionally, to improve the reconstruction quality, LiDAR point clouds can be introduced to supervise the elevation. $L^{*}$ indicates an optional loss.}
  \label{fig:overview}
\end{figure*}

\subsection{Classical 3D Reconstruction}
Reconstructing 3D scenes from multi-view images is a fundamental computer vision task. Traditional 3D reconstruction methods represent scenes as point clouds or meshes, with numerous notable achievements in this area~\cite{ozyecsil2017survey}. COLMAP~\cite{Schonberger2016structure} is a representative of incremental Structure-from-Motion (SfM) methods~\cite{snavely2006photo, snavely2011scene, wu2013towards, moulon2013adaptive}. COLMAP~\cite{Schonberger2016structure} reconstructs 3D scenes by extracting feature points from images and performing feature matching, triangulation, and bundle adjustment. It is capable of reconstructing large-scale scene point clouds and recovering camera poses. However, the reconstructed point clouds tend to be sparse~\cite{liu2023regformer}. Large-scale Multi-View Stereo (MVS) methods~\cite{schonberger2016pixelwise, zhou2021dp, agarwal2011building} can achieve denser scene reconstruction, but they still rely on feature extraction and matching from images, which results in significant holes in the reconstruction of areas with weak textures, such as road surfaces.

\subsection{Implicit 3D Reconstruction}

The emergence of Neural Radiance Fields (NeRF)~\cite{mildenhall2021nerf} has sparked a research wave in implicit scene modeling. NeRF~\cite{mildenhall2021nerf} utilizes Multi-Layer Perceptron (MLP) to learn the mapping from spatial coordinates to scene geometry and colors, achieving continuous space implicit modeling with realistic rendering effects. Mip-NeRF~\cite{barron2021mip} replaces the point-based ray sampling with a multi-variate Gaussian-based pyramidal sampling to achieve anti-aliasing scene modeling. 
Some studies use a combination of grid and MLP for scene modeling~\cite{wang2023co, wu2024dvn}. For large-scale scenes~\cite{liu2024dvlo}, Mip-NeRF 360~\cite{barron2022mip} applies the classic extended Kalman filter concept to compact the scene model of Mip-NeRF~\cite{barron2021mip}, enabling unbounded scene modeling. BungeeNeRF~\cite{xiangli2022bungeenerf} adds scene details at different scales as residual blocks to the neural network, achieving multi-scale modeling of scene details from satellite-scale to fine-scale. There are also a lot of studies on modeling the urban scene~\cite{tancik2022block, turki2023suds, xie2023s}. However, training and rendering are too time-consuming. The emergence of 3D Gaussian Splatting (3DGS)~\cite{kerbl20233d, chen2024survey} greatly improves rendering speed and performance which is widely used in many tasks~\cite{zhu2024semgauss, jiang2024neurogauss4dpci}. However, 3DGS mainly focuses on new perspective synthesis instead of scene geometry recovery. Although some methods use implicit representation to reconstruct 3D mesh~\cite{yu2022monosdf, guedon2023sugar}, they are not yet suitable for large urban scenes.

\subsection{Road Surface Reconstruction}
Algorithms directly targeting road reconstruction can generate dense road surface results~\cite{fan2018road, yu20073d, brunken2020road, guo2015automatic, fan2021rethinking,zhao2024roadbev}, but are limited to the reconstruction of small-scale road areas. For large-scale road surface reconstruction, RoMe~\cite{mei2024rome} utilizes a mesh representation of the road surface and performs rendering optimization to determine the color and semantics of vertexes. EMIE-MAP~\cite{wu2024emie} introduces implicit color encoding to address the issue of inconsistent luminosity in surround-view cameras. However, this mesh-based rendering optimization approach is slow, and the reconstruction quality and efficiency may not meet the requirements of practical applications. 
\section{Approach}
\label{approach}

% \begin{figure}
%   \centering
%   \begin{minipage}{0.49\linewidth}
% 	\centering
% 	\includegraphics[width=0.9\linewidth]{fig/grid1.pdf}
% 	\caption{chutian1}
% 	\label{grid1}%文中引用该图片代号
% \end{minipage}
%   \begin{minipage}{0.49\linewidth}
% 	\centering
% 	\includegraphics[width=0.9\linewidth]{fig/grid2.pdf}
% 	\caption{chutian1}
% 	\label{chutian1}%文中引用该图片代号
% \end{minipage}
% \caption{In this section, we introduce the modeling of road surface using 2D Gaussian surfels.}

% \end{figure}

% \begin{figure}
%   \centering
%     \begin{subfigure}{0.5\linewidth}
%       \centering   
%       \includegraphics[width=1\linewidth]{fig/grid1.pdf}
%         \caption{caption\_for\_sub1}
%         \label{fig:sub1}
%     \end{subfigure} \begin{subfigure}{0.5\linewidth}
%       \centering   
%       \includegraphics[width=\linewidth]{fig/grid2.pdf}
%         \caption{caption\_for\_sub2}
%         \label{fig:sub2}
%     \end{subfigure}
% \caption{
% \label{fig:total}
% write\_caption\_here
% }
% \end{figure}

The overview of RoGs is shown \cref{fig:overview}, which consists of three main components. First, we use meshgrid Gaussian to represent the real road surface (\cref{sec:surface}). The Gaussian surfels store position, scale, rotation, color, opacity, and semantic information. Then, for each Gaussian surfel, its elevation (z) and rotation are initialized using the nearest vehicle pose (\cref{sec:init}). Finally, the rendered RGB and semantics are supervised by the camera images and semantic segmentation results. (\cref{sec:opt}). Additionally, LiDAR point clouds can be introduced to supervise the elevation of surfels to improve the reconstruction quality.

\subsection{Representation with Meshgrid Guassian}
\label{sec:surface}
\begin{figure}[htbp]
  \centering
    \begin{subfigure}{0.25\linewidth} 
      \includegraphics[width=\linewidth]{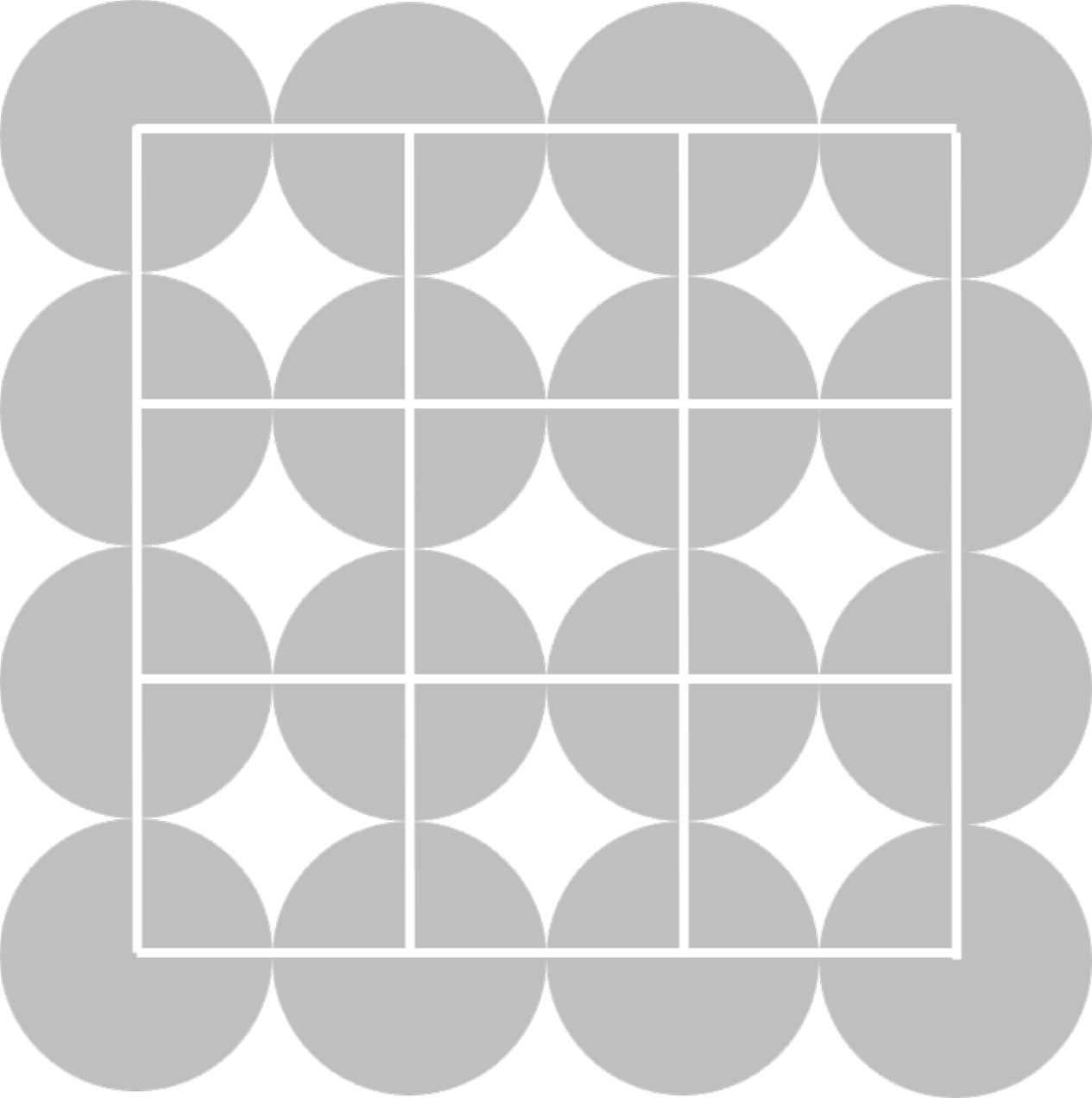}
        \caption{Layout-1}
        \label{fig:layout1}
    \end{subfigure}  \qquad \begin{subfigure}{0.27\linewidth}
      \includegraphics[width=\linewidth]{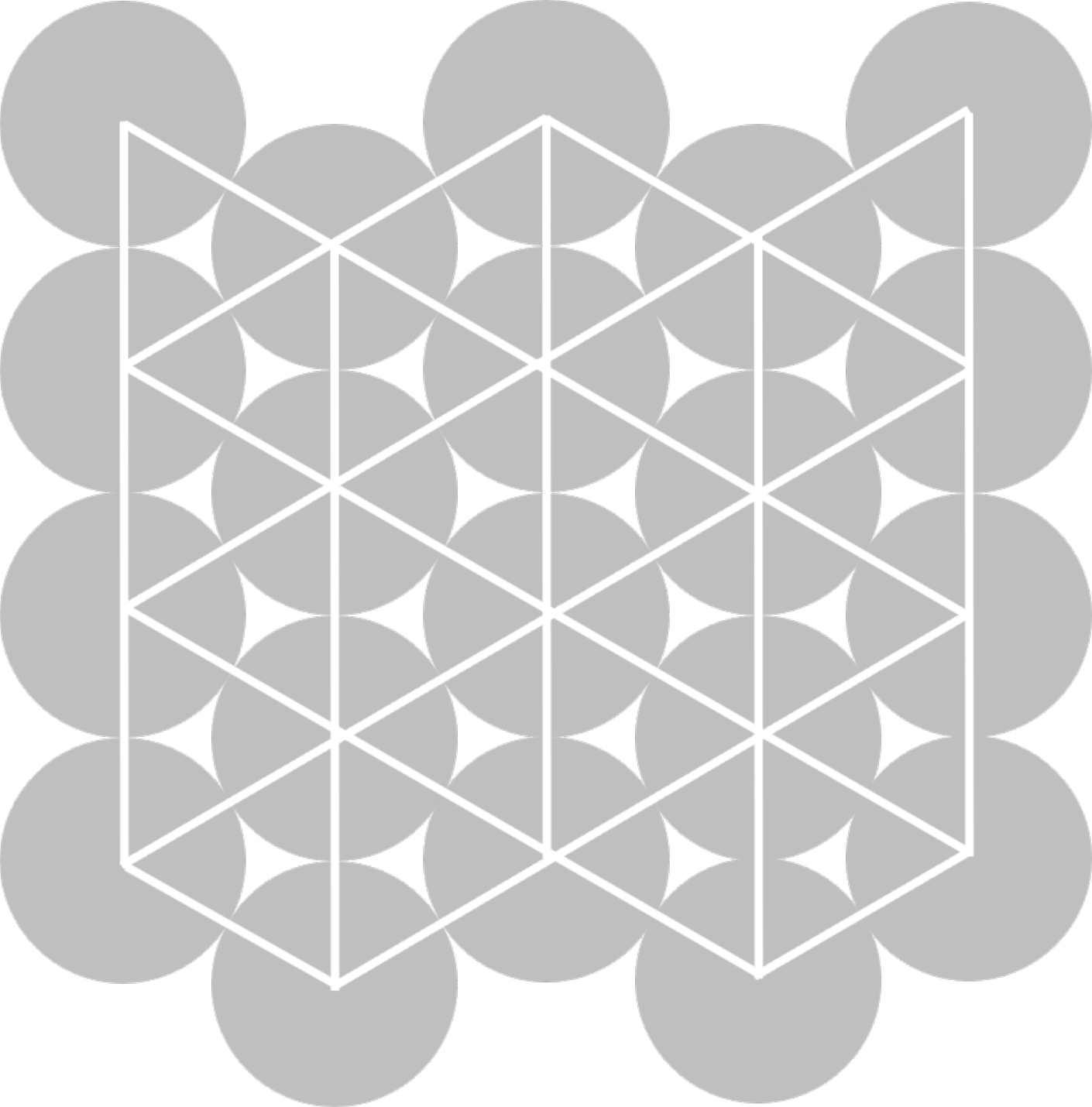}
        \caption{Layout-2}
        \label{fig:layout2}
    \end{subfigure}
\caption{Comparison of the two layouts. Layout-1 covers the entire road with fewer Gaussian surfels and reduces the overlap between Gaussian surfels during training. Therefore, Layout-1 is used to place Gaussian surfels.}
\label{fig:meshgrid}
\end{figure}

In this section, we introduce the modeling of road surface using Meshgrid Gaussian. 

\textbf{2D Gaussian Surfel:} In 3DGS~\cite{kerbl20233d}, a Gaussian sphere in the world space is modeled by a 3D covariance matrix $\mathbf{\Sigma}$ and its center coordinates $p$:
\begin{equation}
  G(p)\:=e^{-\frac{1}{2}p^T\mathbf{\Sigma}^{-1}p},
\end{equation}
where the 3D covariance matrix $\mathbf{\Sigma}$ is determined by the rotation $\mathbf{R}$ and scale $\mathbf{s}=(s_x, s_y, s_z)^T$ along three dimensions. For convenience, the scale is represented as a diagonal matrix $S=Diag(s_x, s_y, s_z)$. The formula for obtaining $\Sigma$ is:
\begin{equation}
  \mathbf{\Sigma} = \mathbf{RSS}^T\mathbf{R}^T.
\end{equation}

Similar to previous work~\cite{dai2024high, huang20242d}, our method simply sets the $s_z$ of the Gaussian sphere to 0 to obtain a Gaussian surfel. Compared to Gaussian spheres, the Gaussian surfels representation aligns more closely with the physical reality of the road. To represent texture information and semantic information, we parameterize the color (RGB), opacity, and semantics of the Gaussian surfel. Ultimately, a Gaussian surfel can be parameterized explicitly as:
\begin{equation}
  \Theta = \{(x,y,z), (r,g,b), (s_x, s_y), \alpha, \mathbf{R}, Sem\},
\end{equation}
where $(x,y,z)$ represents the center coordinates, $(r,g,b)$ represents the color, $s_x$ and $s_y$ represent the scales in the $x$ and $y$ directions, $\alpha$ represents opacity, $\mathbf{R}$ represents rotation, and $Sem$ represents semantics. In practice, we use a vector of length equal to the number of semantic categories to represent semantics.

\textbf{Meshgrid Layout:} In order to cover the entire road surface, the meshgrid layout shown in \cref{fig:layout1} is used to place Gaussian surfels. First, the vehicle poses are projected onto the xy plane, and the road range is further expanded. Then, it is divided into an evenly distributed square grid, as shown in \cref{fig:layout1}, with Gaussian surfels placed at the vertices of the grid. Compared to the Layout-2 shown in \cref{fig:layout2}, C, Layout-1 can cover the road with fewer Gaussian surfels. In addition, since the size of Gaussian surfels is learnable, overlap between neighboring surfaces occurs with training. Larger overlaps interfere with the optimization, whereas the Layout-1 is able to reduce the overlap between Gaussian surfels because of fewer neighbors.

\begin{figure}[tbp]
  \centering
  \includegraphics[width=\linewidth]{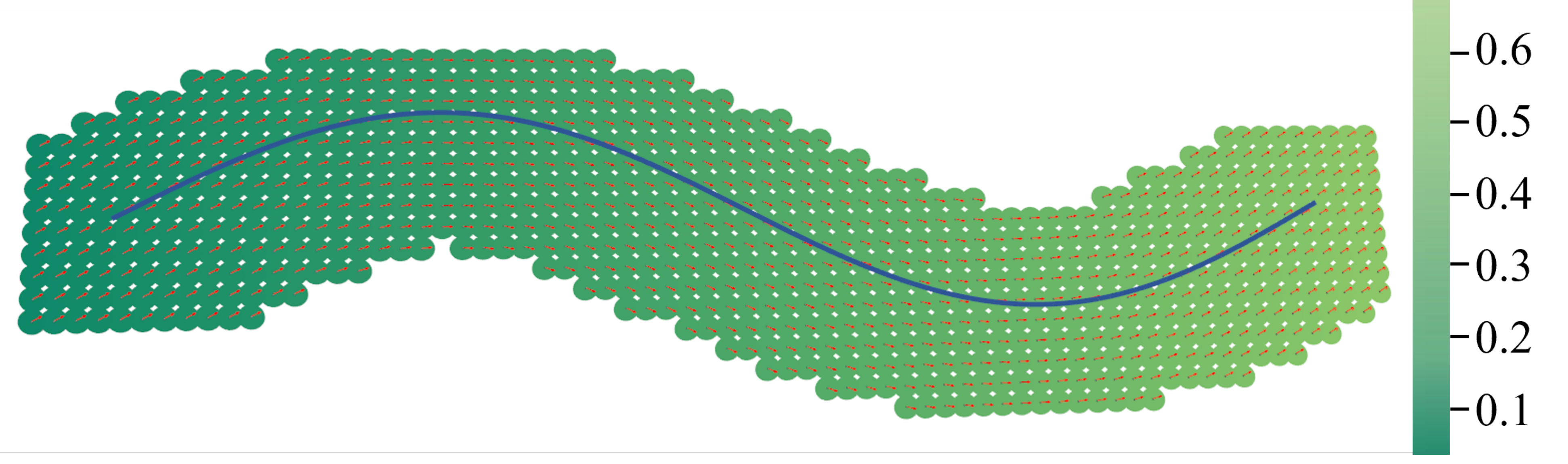}
  \caption{Pose-based initialization. For each Gaussian surfel, its elevation and rotation are initialized using the nearest vehicle pose. The blue curve represents the vehicle trajectory, and the red arrows represent the x-axis.}
  \label{fig:road}
\end{figure}

\subsection{Pose-based Initialization}
\label{sec:init}

After obtaining the initial Gaussian surfels coordinates by projecting the vehicle poses onto the xy-plane, it is not aligned with the road surface in terms of elevation because z is set to 0. Considering that vehicle poses are commonly parallel to the road, for a Gaussian surfel, its z-coordinate is initialized using the height and rotation of the nearest vehicle pose, and its rotation is initialized to the pose of the nearest vehicle rotation. For example, a Gaussian surfel located at $(x_i,y_i)$ whose corresponding to the nearest vehicle position and rotation denoted as $(x_v,y_v,z_v)$ and $\mathbf{R}_v=(\mathbf{n}_1, \mathbf{n}_2, \mathbf{n}_3)$, respectively, where $\mathbf{n}_3=(n_{31},n_{32},n_{33})^T$. Then the rotation of the Gaussian surfel is initialized to $\mathbf{R}_v$, and the $z_i$ is initialized using the following equation:
\begin{equation}
  z_i = z_v-\frac{1}{n_{33}}(n_{31}\cdot(x_i-x_v)+n_{32}\cdot(y_i-y_v)).
\end{equation}
In practice, $z_i$ can be approximated to be initialized using $z_v$ when the roll angle of the vehicle pose is small.

The initialization result is shown in \cref{fig:road}. This initialization method fully leverages the relationship between the vehicle and the road surface, which is more conducive to the subsequent optimization of road surface geometric textures. 

\begin{figure}[ht]
  \centering
  \includegraphics[width=\linewidth]{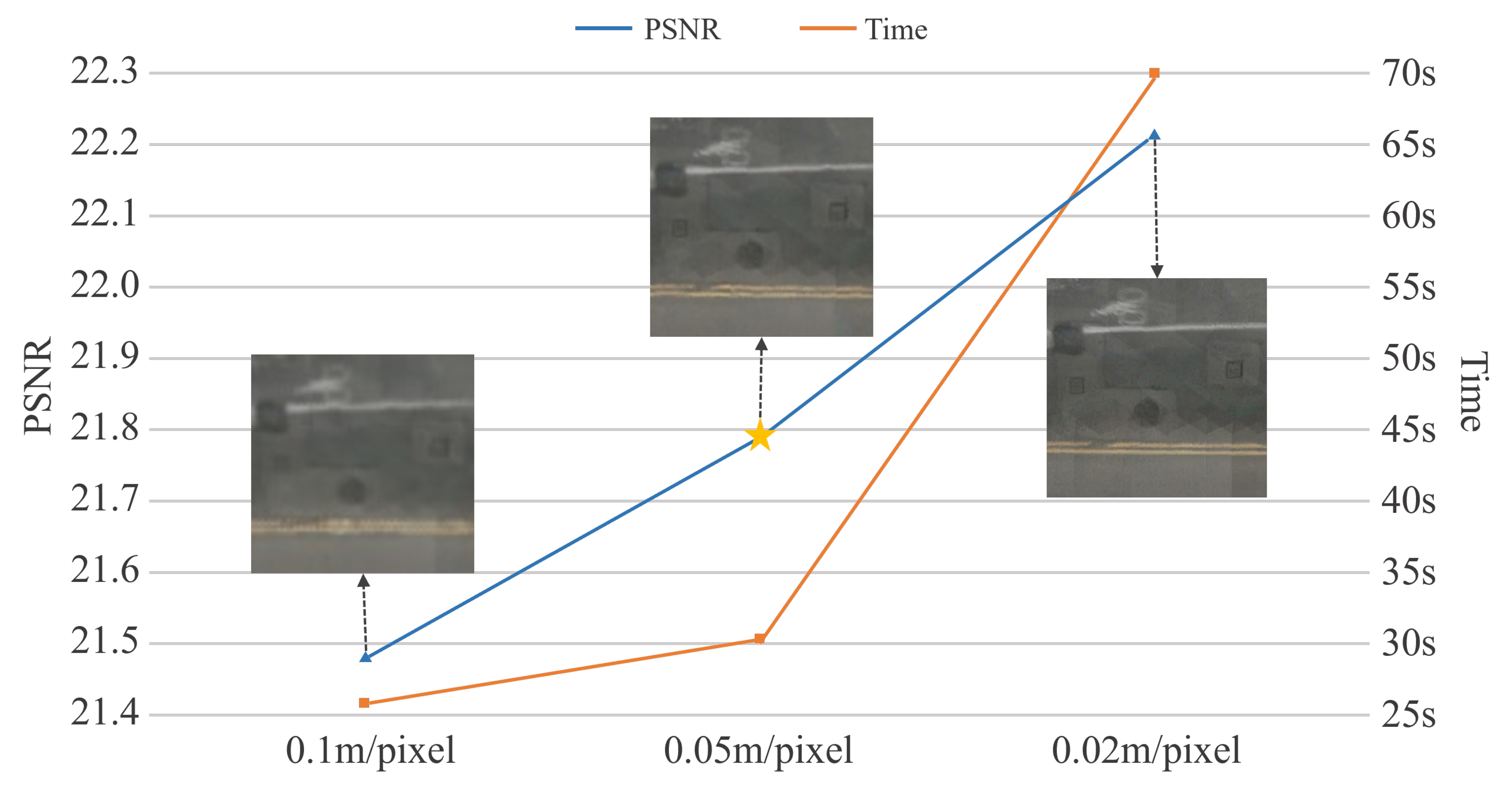}
  \caption{Results at different resolutions. 0.05m/pixel resolution achieves realistic reconstruction with improved training speed.}
  \label{fig:resolution}
\end{figure}

\begin{figure*}[t]
  \centering
  \includegraphics[width=\linewidth]{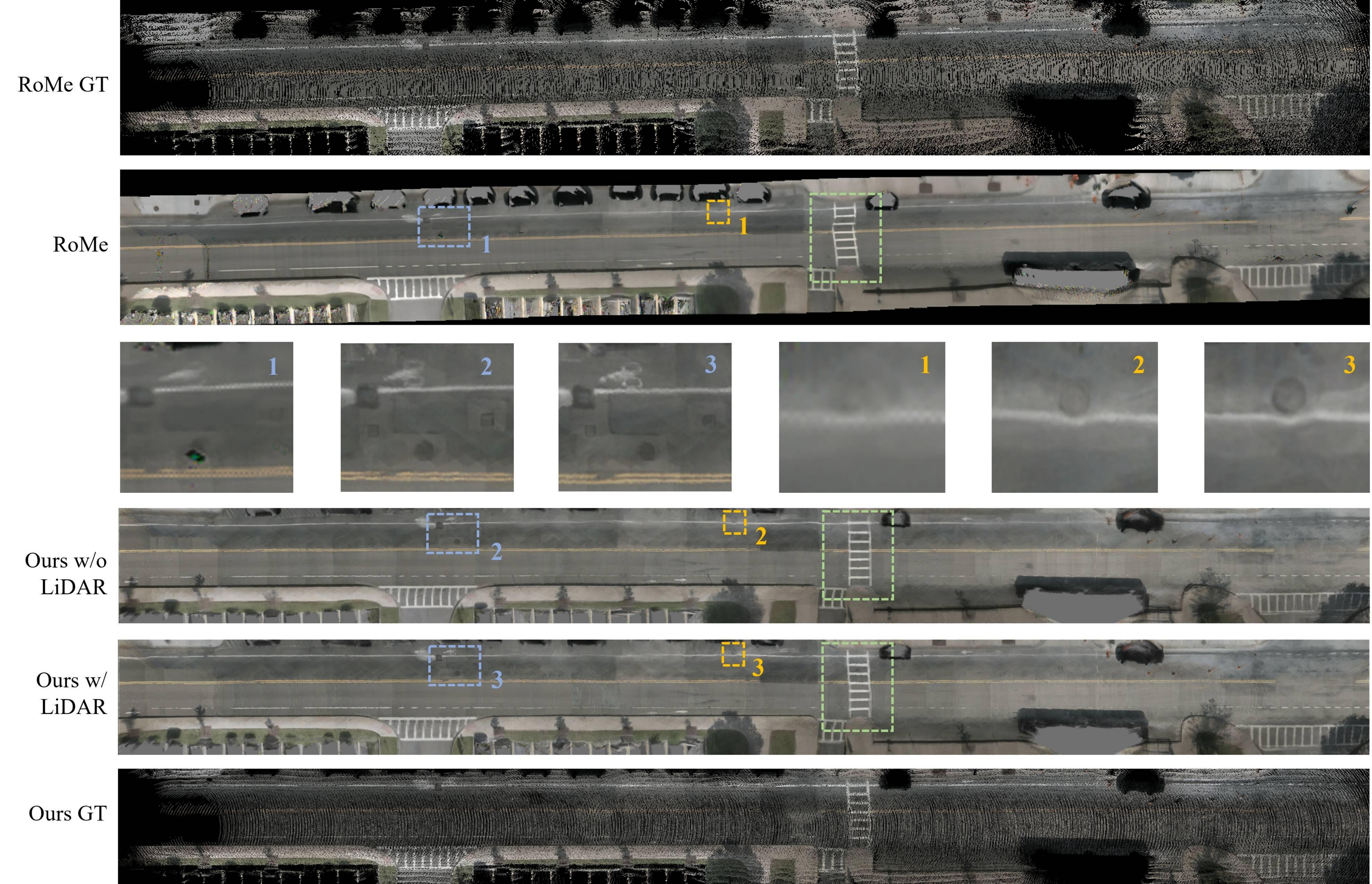}
  \caption{RGB results of road surface reconstruction for Scene-0655. Because RoMe is in a different reference to ours, the ground truth needs to be transformed into the corresponding reference for rendering. In the local detail images, our method reconstructs both the ground bike markings and lane lines more clearly, which shows that our method is able to reconstruct finer details. In addition, when height is supervised using LiDAR, the quality and detail of the rendering are further improved.}
  \label{fig:0655_rgb}
\end{figure*}

\subsection{Optimization}
\label{sec:opt}

\subsubsection{RGB and Semantic Rendering}
\label{sec:render}

To render the image, the Gaussian surfels are first transformed from world coordinates to camera coordinates using the world-to-camera transformation matrix $W$, and then projected onto the image plane through a local affine transformation $J$  to obtain a new covariance matrix $\mathbf{\Sigma}^{\prime}$:
\begin{equation}
  \mathbf{\Sigma}^{\prime} = JW\mathbf{\Sigma}W^TJ^T.
\end{equation}
In particular, if the image is rendered using an orthogonal camera, we only need to set $J$ to a unit matrix. 

Since $\mathbf{\Sigma}^{\prime}$ is projected onto a 2D plane, a 2D covariance can be obtained by skipping the third row and column, forming an ellipse $g$ on the image plane. For a pixel in the image, after sorting all $K$ projected $g$ onto it in ascending order of original depth, the color of that point can be represented as:
\begin{equation}
\mathbf{c}(p)=\sum\limits_{k=1}^K\mathbf{c}_k \alpha_k g_k(p)\prod\limits_{i=1}^{k-1}(1-\alpha_i g_i(p)),
\end{equation}
where $c$ represents color and $\alpha$ represents opacity. When rendering semantics, the color $c$ is replaced by the semantic vector. Due to the large number of Gaussian surfels on the ground of a large scene, in order to speed up rendering, only Gaussian surfels within a certain range in front of the vehicle camera are used for rendering. Details are provided in the supplementary material.

Additionally, in the real world, each camera has a different level of exposure. Therefore, when reconstructing the road surface using multiple cameras, we introduce a learnable exposure parameter $a$ and $b$ for each camera. In this case, the final output color is as follows:
\begin{equation}
\mathbf{c}^{\prime}(p)= e^a \cdot \mathbf{c}(p) +b.
\end{equation}

\begin{figure*}[tb]
  \centering
  \includegraphics[width=\linewidth]{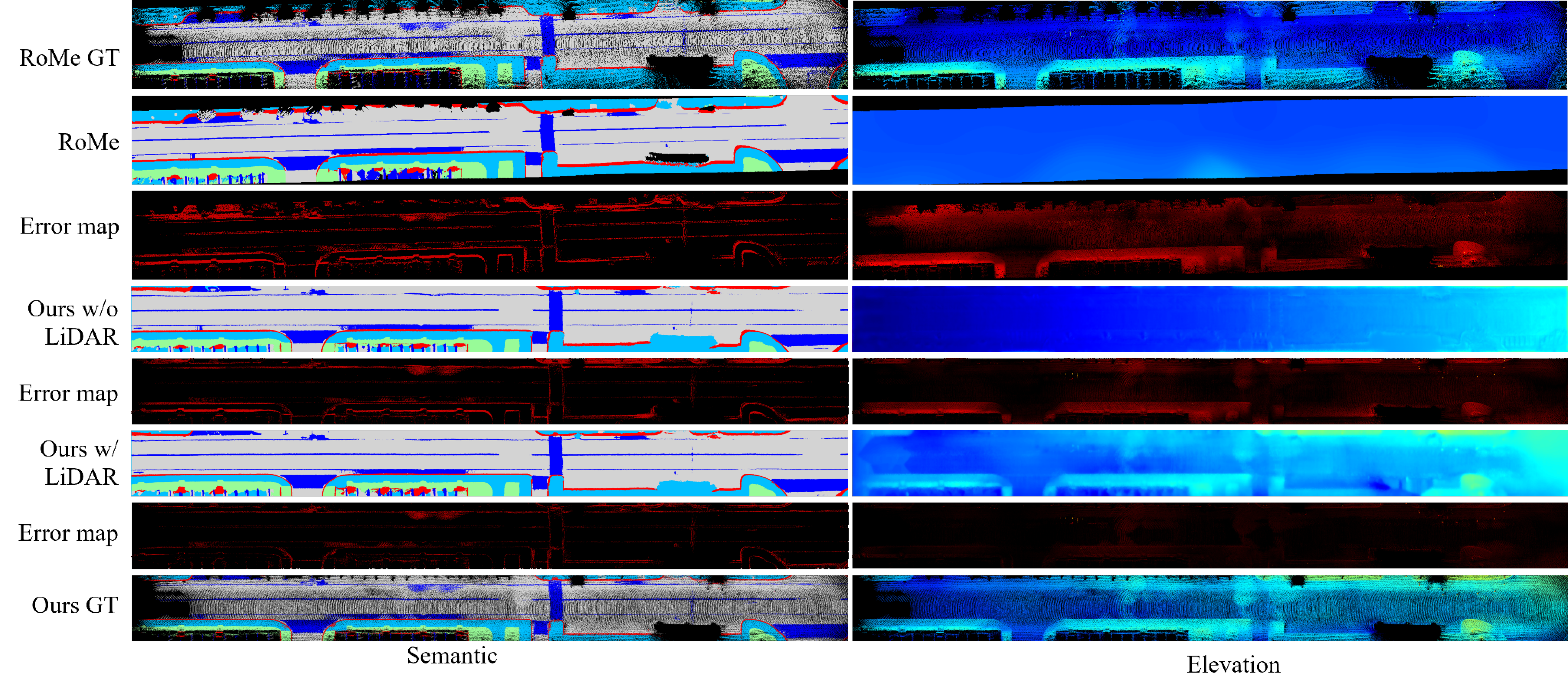}
  \caption{Semantic and elevation results of road surface reconstruction for Scene-0655. For the elevation error map, brighter colors indicate larger errors. For the semantic error map, incorrect pixels are shown in red. The error maps show that our method has fewer errors. In addition, when neither LiDAR point cloud is used, the reconstructed elevation of our method restores the height of road edges and zebra crossings. In contrast, the elevation reconstructed by RoMe is too smooth and does not correspond to real-world physical road surfaces.}
  \label{fig:0655_sem}
\end{figure*}

\subsubsection{Loss}
\label{sec:loss}
For color, the rendered images are supervised by the camera images. For semantics, the rendered semantics are supervised using the semantic results inferred by Mask2Former~\cite{cheng2022masked}. A semantic-based mask $M$ is used to remove non-road elements. The color loss and semantic loss are as follows:
\begin{align}
L_{c}&=\frac1{|M|}\sum m_i \cdot |c_i-\bar{c_{i}}|,\\
L_{s}&=\frac1{|M|}\sum m_i \cdot CE(sem_i,\bar{sem}_i),
\end{align}
where $\bar{c_i}$ and $\bar{sem}_i$ represent the color ground truth and semantic ground truth, respectively. $m_i$ is a mask value that is either 0 or 1. $CE(\cdot)$ represents the cross-entropy loss. $|M|$ represents the number of valid pixels in the mask $M$. 

Due to the smoothness of the road surface, we introduce an elevation smoothness loss as follows:
\begin{equation}
L_{smooth}=\frac{1}{K}\sum_{i=1}^N\sum_{j\in N(i)}\|z_i-z_j\|^2_2,
\end{equation}
where $N(i)$ represents the $K$ nearest neighbors. 

It is important to note that we do not use the approach of finding $K$ nearest neighbors by calculating the distance and then sorting. This is because the number of Gaussian surfels in a large scene is huge and it would consume a lot of time. To solve this case, we assign Gaussian surfels to image pixel indexes in the $xy$-plane, and compute the $K$ nearest neighbors efficiently by querying the pixel indexes. Details are provided in the supplementary material.

% Additionally, to improve the reconstruction results, LiDAR point clouds can be introduced to supervise the elevation:
Additionally, to improve the reconstruction results, LiDAR point clouds can be used for elevation supervision:
\begin{equation}
L_z=\sum_{i=1}^N \|z_i-\bar{z_i}\|^2_2,
\end{equation}
where $\bar{z_i}$ is obtained by querying the nearest neighbor elevation of the point cloud in the xy-plane.

The total loss is the weighted sum of these losses:
\begin{equation}
L_{total}=\lambda_{c}L_{c}+\lambda_{s}L_{s}+\lambda_{smooth}L_{smooth}+\lambda_zL_z,
\end{equation}
where $\lambda_{c}$, $\lambda_{s}$,$\lambda_{smooth}$ and $\lambda_{z}$ represent the corresponding weights.

\section{Experiments}
\begin{table*}[htbp]
  \caption{Results on the nuScenes dataset. The mIoU is expressed as a percentage. ``Elev.'' denotes the RMSE of the elevation error with units in $m$. ``Ep.1'' and ``Ep.2'' denote training one epoch and two epochs respectively. When neither LiDAR point cloud is used, the PSNR of our method is the same as RoMe~\cite{mei2024rome}, and it is improved in both mIoU and Elev. metrics that reflect the 3D structure.  Importantly, we achieve a $\mathbf{53\times}$ speedup. In addition, it can be noticed that there is a significant improvement in all metrics when using LIDAR.}
  \label{tab:metric}
  \centering
  \resizebox{\linewidth}{!}{
  \begin{tabular}{l||ccc|ccc|ccc|ccc|ccc|cccl}
    \toprule
    \multirow{2}{*}{Method} &\multicolumn{3}{c|}{scene-0064} & \multicolumn{3}{c|}{scene-0212}& \multicolumn{3}{c|}{scene-0523}& \multicolumn{3}{c|}{scene-0655}&\multicolumn{3}{c|}{scene-0856}&\multicolumn{4}{c}{Mean}\\
    \cline{2-20}
    & PSNR & mIoU& Elev. & PSNR & mIoU& Elev.& PSNR & mIoU& Elev.& PSNR & mIoU& Elev.& PSNR & mIoU& Elev. & PSNR & mIoU& Elev. &Time\\
    \hline
    \hline
    RoMe~\cite{mei2024rome} & 23.38  &84.48 &0.108 & 25.58&89.11&0.180&25.83&86.37&0.118&20.56&84.24&0.164&22.79&80.53&0.311&23.63&84.95&0.176&1630.0~\\
    
    Ours w/o LiDAR Ep.1 & 23.16  &83.21 &0.131 & 25.41&91.80&0.115&24.61&87.07&0.128&21.79&86.37&0.107&23.20&77.85&0.290&23.63&85.26&0.154&30.7 ~\textcolor{red}{\textbf{53$\times$}}\\
    Ours w/o LiDAR Ep.2& 23.00  &83.38 &0.129 & 24.99&91.81&0.115&24.66&87.08&0.128&21.66&86.49&0.107&22.91&77.94&0.290&23.44&85.34&0.154&61.1~\textcolor{red}{\textbf{27$\times$}}\\
    \midrule
    Ours w/ LiDAR Ep.1 & 24.38  &91.91 &0.060 & 25.85&93.70&0.073&25.57&91.53&0.058&22.92&92.34&0.066&23.27&82.56&0.229&\textbf{24.40}&90.41&0.097&32.1~\textcolor{red}{\textbf{51$\times$}}\\
    Ours w/ LiDAR Ep.2& 24.24 &93.73 &0.058 & 25.55&93.95&0.070&25.34&92.38&0.054&23.10&93.21&0.065&23.09&84.34&0.222&24.26&\textbf{91.52}&\textbf{0.094}&63.9~\textcolor{red}{\textbf{25$\times$}}\\
    \bottomrule
  \end{tabular}
  }
\end{table*}

\subsection{Dataset}
\textbf{NuScenes:} We conduct experiments on the NuScenes~\cite{caesar2020nuscenes} dataset, which consists of 1000 video scenes, each lasting 20 seconds. The vehicle is equipped with 6 cameras and one LiDAR. The frequencies of the cameras and LiDAR are 12Hz and 20Hz, respectively. We use the 6 surround-view images and LiDAR for road surface reconstruction. We test on some scenes from the NuScenes dataset. 

\textbf{KITTI:} The KITTI~\cite{geiger2013vision} dataset consists of 22 sequences, with sequence 00 covering an area of approximately 600 m $\times$ 600 m. We conduct experiments on this larger-scale scene. The KITTI collection vehicle is equipped with two color cameras, but we only used the left color camera. 

For both datasets, a pre-trained Mask2Former~\cite{cheng2022masked} model is used to obtain semantic results. To ensure fairness, we use the same pre-trained model as RoMe~\cite{mei2024rome} to infer to get semantic segmentation results.

\subsection{Experiment Setup}
\label{detail}
\textbf{Metric:} Since there is no Bird's Eye View (BEV) ground truth available, the ground truth is constructed using concatenated LiDAR point clouds. Each frame of the point cloud is projected onto the 6 images of the nearest frame, where the color and semantic labels of each point are determined by the projected points on the images. Points that do not belong to the road surface elements are removed. For points visible in the overlapping areas of multiple cameras, the color and semantics take the mean and the mode, respectively. 

The concatenated dense LiDAR point cloud is rendered as an RGB image and a semantic map under BEV, which is used to evaluate the color and semantics of the reconstruction results. We use the Peak Signal-to-Noise Ratio (PSNR) to evaluate the image quality and mean Intersection over Union (mIoU) to evaluate the accuracy of semantics. 

For the evaluation of the reconstructed 3D structure, we use the elevation of the concatenated  point cloud. Specifically, for the reconstructed center point, the elevation of the nearest point in the point cloud is queried for a certain radius (0.1m) in the $xy$-plane as the ground truth to calculate the error. The RMSE of the elevation error is used as the elevation metric. 

\textbf{Implementation Details:} Our experiments are run on a Linux server with an RTX-4090. The learning rates for $\alpha$, $(s_x,s_y)$, and $\mathbf{R}$ are all 1e-4. The learning rate of $z$ is proportional to the scene size, and this ratio is initially 1.6e-4 and eventually decreases to 1.6e-6. The learning rates for color and semantics are 0.008 and 0.1, respectively. The learning rate for the exposure parameter is 0.001. The optimizer is Adam. The loss weights are set as  $\lambda_{c}$=1, $\lambda_{s}$=0.06, $\lambda_{smooth}$=0.003 and $\lambda_{z}$=0.02. In particular, $\lambda_{smooth}$ is adjusted to 1 when the LiDAR point cloud is used to supervise the height. 

In order to trade-off between training speed and reconstruction quality, we conduct experiments on BEV resolution using scene-0655 from the nuScenes dataset. The results are shown in \cref{fig:resolution}. BEV resolutions greater than or equal to 0.1 m/pixel result in blurred reconstruction while less than or equal to 0.02 m/pixel resolution increases the unnecessary computational. Therefore, 0.05 m/pixel resolution (marked with a star) is the best choice.

\begin{figure}[htbp]
  \centering
    \begin{subfigure}{\linewidth} 
    \centering
      \includegraphics[width=\linewidth]{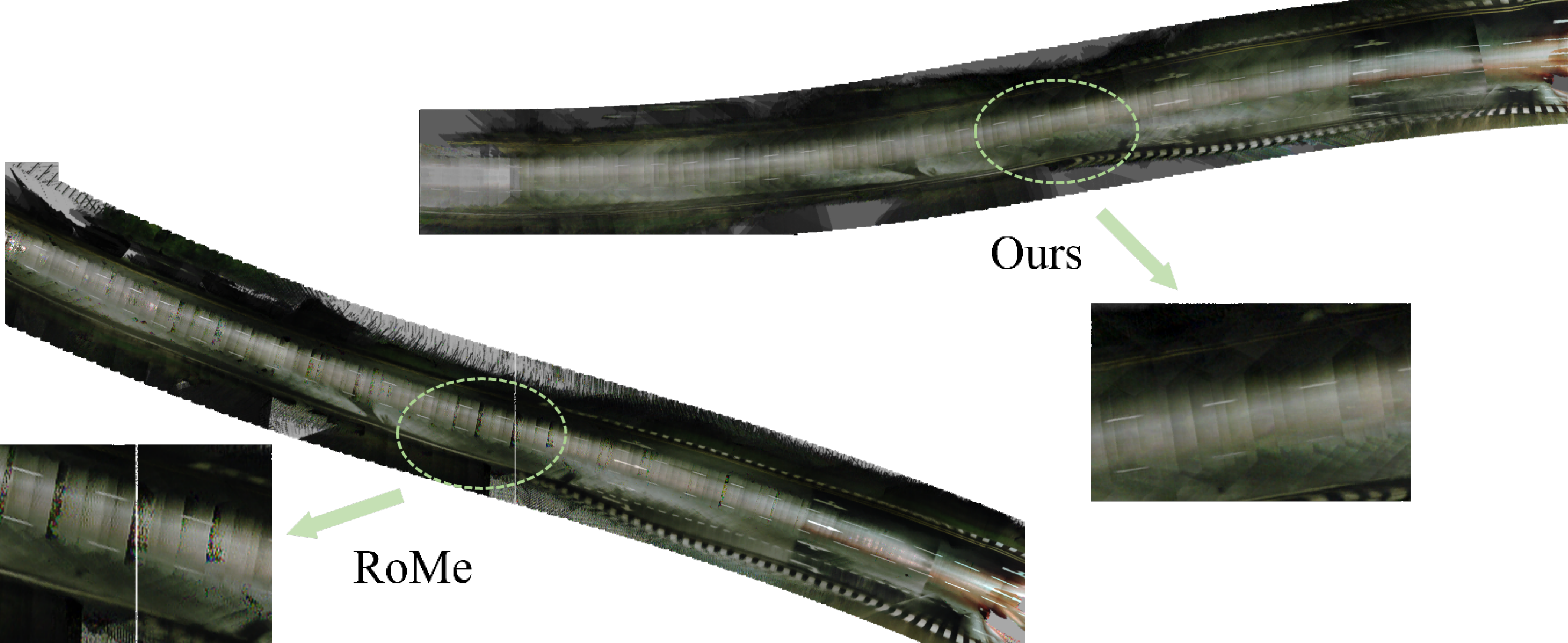}
        \caption{Scene-1079 (Night)}
        \label{fig:1079}
        \vspace{3mm}
    \end{subfigure} 
    \begin{subfigure}{\linewidth}
    \centering
      \includegraphics[width=\linewidth]{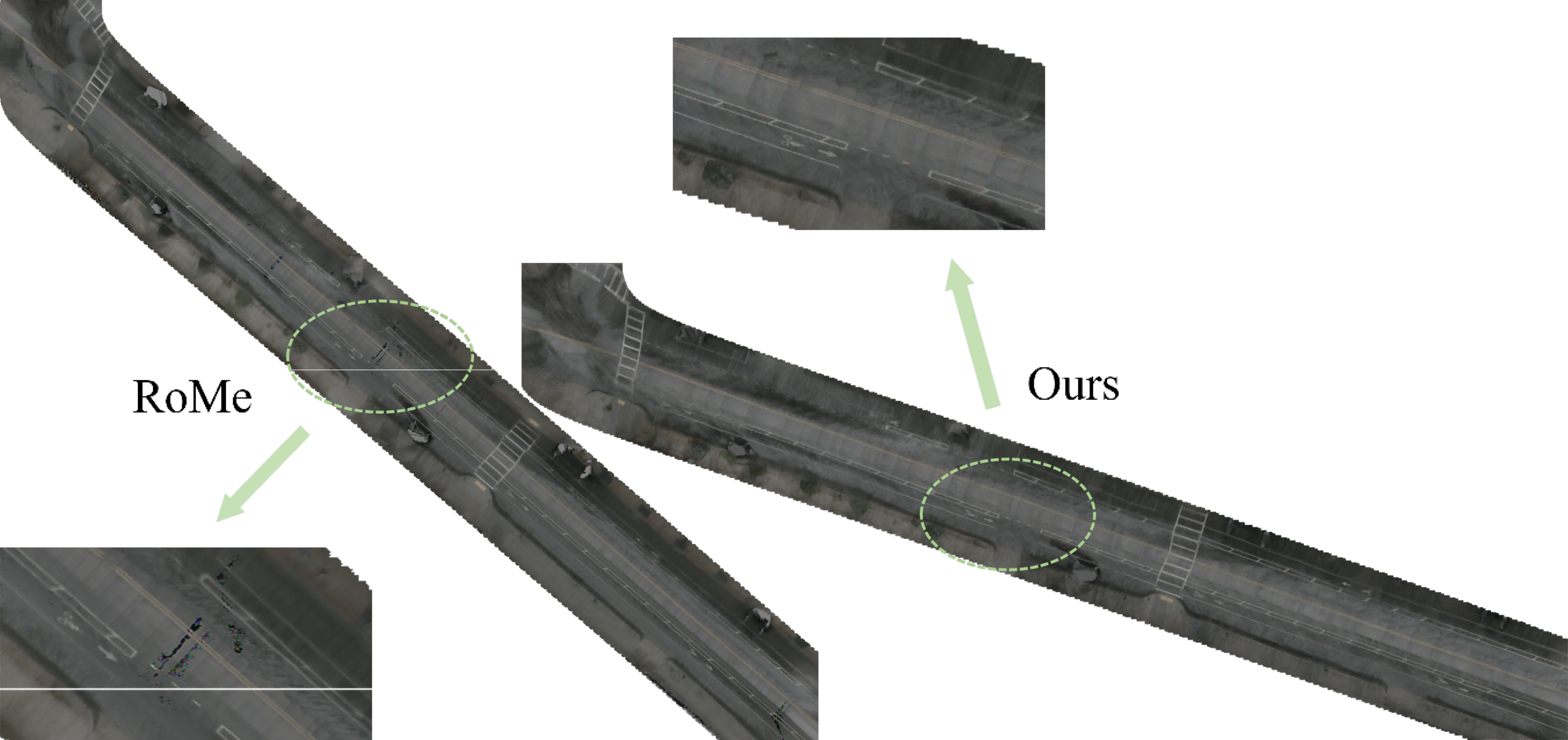}
        \caption{Scene-0580 (Rainy)}
        \label{fig:0580}
    \end{subfigure}
\caption{Comparison of road reconstruction results for night and rainy conditions between RoMe~\cite{mei2024rome} and ours without LiDAR point cloud. In challenging scenes, the reconstruction results of RoMe~\cite{mei2024rome} show obvious RGB noise, and our method is more robust compared to it.}
\label{fig:night_rainy}
\end{figure}

\subsection{Experimental Results}
\label{sec:compare}
The rendering results in a BEV, obtained using an orthographic camera. For large scenes, in order to speed up the rendering, the scene is rendered in chunks and then stitched together. Details are provided in supplementary material.

\textbf{Comparison with RoMe:}
\cref{tab:metric} shows the evaluation results of the road surface reconstruction for the five scenes in nuscenes. In both cases without using LiDAR point clouds, our method reconstructs results with a similar PSNR as RoMe~\cite{mei2024rome}, but there is a significant improvement in mIoU and Elev., which reflect the 3D structure. Importantly, we achieve a $\mathbf{53\times}$ speedup. Additionally, it can be noticed that all metrics are significantly improved when using the LiDAR point cloud. In addition, compared to training one epoch, the results of two epochs decrease slightly in PSNR and get a boost in mIoU, with the height remaining essentially the same. In practical applications, some semantic reconstruction accuracy can be lost for faster speed.

\begin{figure}[tbp]
  \centering
    \begin{subfigure}{\linewidth} 
    \centering
      \includegraphics[width=\linewidth]{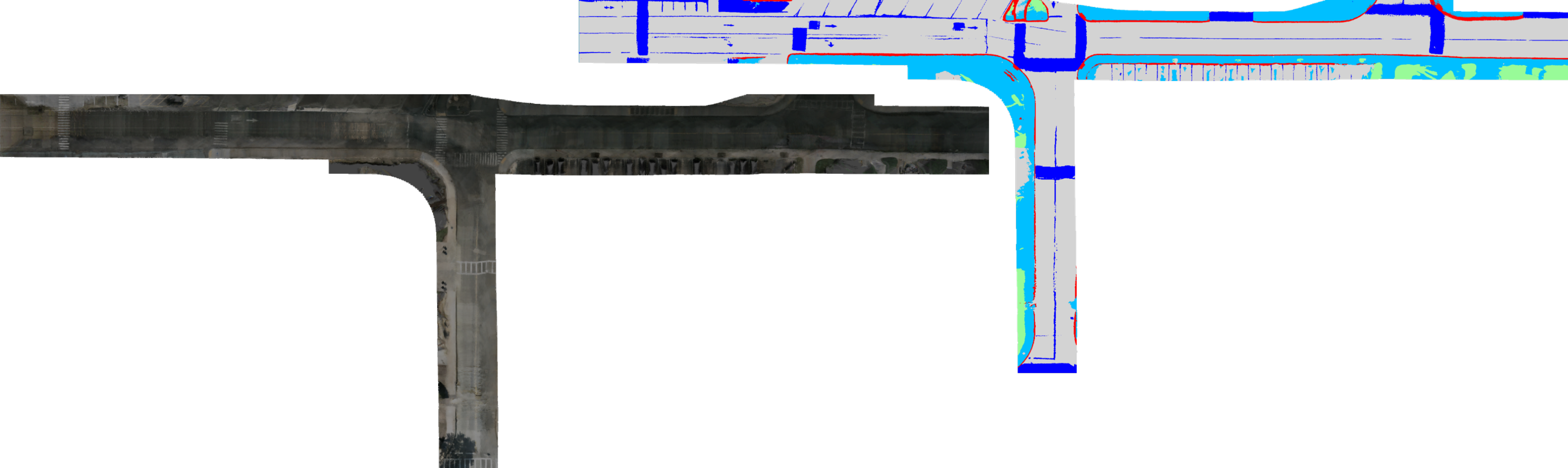}
        \caption{Scene-1}
        \label{fig:scene1}
        \vspace{3mm}
    \end{subfigure} 
    \begin{subfigure}{\linewidth}
    \centering
      \includegraphics[width=\linewidth]{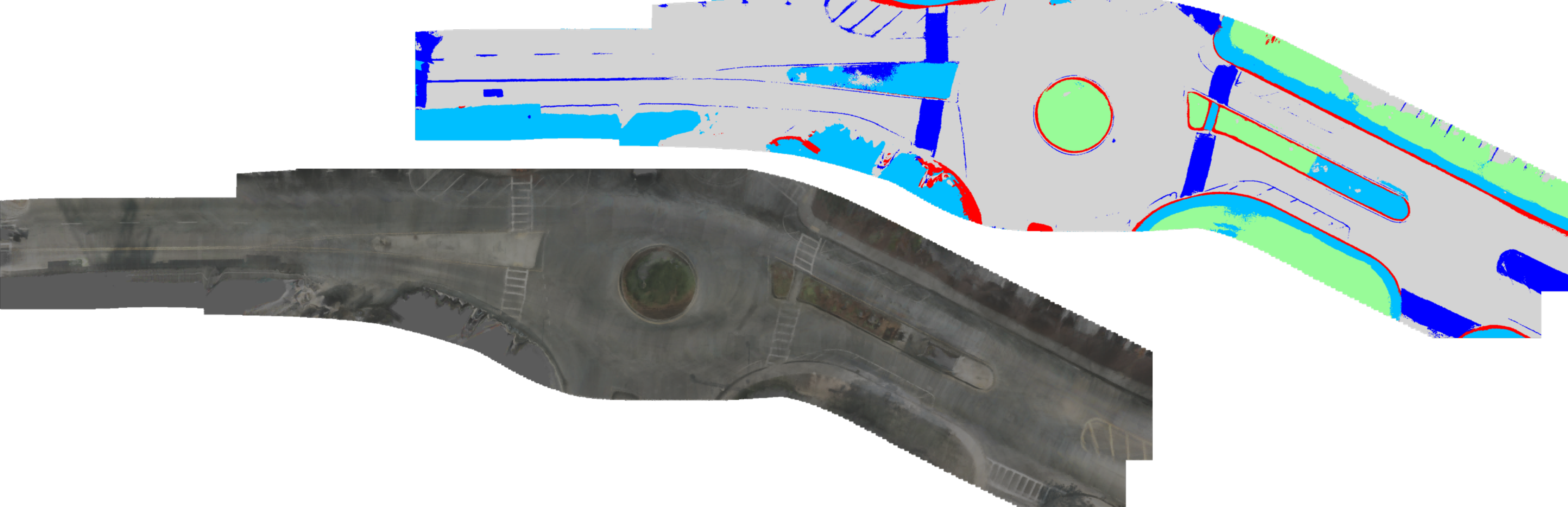}
        \caption{Scene-2}
        \label{fig:scene2}
    \end{subfigure}
\caption{Road surface reconstructions of Scene-1 and Scene-2 with LiDAR point cloud.}
\label{fig:merge_scene}
\end{figure}

\cref{fig:0655_rgb} and \cref{fig:0655_sem} are the visualization results of the RGB map, semantic map and elevation maps of Scene-0655. The error maps show that our method has fewer errors. Without the use of LiDAR point clouds, the reconstructed elevations of our method restore the heights of road edges and zebra crossings to some extent. In contrast, the elevations reconstructed by RoMe~\cite{mei2024rome} are too smooth and do not match the actual road surface in the real world. This is because RoMe~\cite{mei2024rome} uses MLP to model the road surface heights, which is smooth. When using LiDAR, there is only a small error between the reconstructed result of the road surface height and the ground truth.

\cref{fig:night_rainy} shows a comparison of the reconstruction results for the challenging scenes (night and rainy). Both our method and RoMe~\cite{mei2024rome} do not use LiDAR point clouds.  The reconstruction results of RoMe~\cite{mei2024rome} show obvious RGB noise, and our method is more robust compared to it.

\textbf{Multiple Scenes Merging:} \cref{fig:merge_scene} shows the results of merging several scenes. We use the same two merged scenes as in RoMe~\cite{mei2024rome}: Scene-1 and Scene-2, which are composite of four separate scenes. As shown in \cref{fig:merge_scene}, our method gives excellent reconstruction results even for merged scenes consisting of multiple scenes.

\subsection{Ablation}
When using LiDAR point clouds, even without pose-based initialization, the road surface can quickly converge to ground truth through supervised elevation. Therefore, we perform ablation experiments without LiDAR point clouds. In order to verify the effect of the pose-based initialization method, we remove pose-based initialization by setting all $z$ to 0 and rotation to identity matrix. The results are averages obtained by training an epoch in the five scenes of \cref{tab:metric}.  As shown in \cref{tab:ablation}, the experiment results demonstrate the effectiveness of the meshgrid Gaussian and pose-based initialization.

\begin{table}[t]
  \caption{Ablation study results.}
  \label{tab:ablation}
  \centering
  \begin{tabular}{lll|ccc}
    \toprule
     Layout-1 & Layout-2 & Init & PSNR & mIoU & Elev. \\
    \midrule
       \checkmark & & &23.13    &70.28  &0.966   \\
      & \checkmark&  \checkmark& 23.38   &\textbf{85.26} &0.159   \\
     \checkmark &  & \checkmark &\textbf{23.63}    &\textbf{85.26}  &\textbf{0.154} \\
    \bottomrule
  \end{tabular}
\end{table}

\subsection{Limitations}
\label{limit}
Because the reconstruction of the road surface relies on the accuracy of the pose, it often requires more precise vehicle poses in practical use. Although reconstruction results can be fine-tuned by optimizing road colors and elevations, the capacity for this adjustment is limited.  Furthermore, due to the sparse texture information on the road surface, relying solely on ground color and semantics to accurately restore road height can be quite challenging. This issue is more pronounced in RoMe~\cite{mei2024rome}. Therefore, introducing point clouds for elevation supervision is a highly effective method. For vehicles without equipped LiDAR, point clouds obtained from COLMAP~\cite{Schonberger2016structure} can be used as an alternative.
\section{Conclusion}
We propose a novel method for large-scale road surface reconstruction with meshgrid Gaussian, which utilizes mesh-distributed Gaussian surfels to achieve efficient and high-quality road surface reconstruction. The Gaussian surfels are placed on uniformly distributed square mesh vertices, this layout covers the entire road with fewer Gaussian surfels and reduces the overlap between surfels. In addition, we introduce a pose-based initialization method. Given the prior condition that vehicle poses are parallel to the road surface, the elevation and rotation of Gaussian surfel are initialized using the nearest vehicle pose. This initialization method fully leverages the relationship between vehicle trajectories and the road surface, which facilitates subsequent optimization of road geometry and texture. Our method achieves outstanding results in various challenging real-world scenes. Notably, the reconstruction process is completed within a remarkably short timeframe.

% More importantly, we can complete the reconstruction in a very short time. 

{
    \small
    \bibliographystyle{ieeenat_fullname}
    \bibliography{main}
}

\clearpage
\setcounter{page}{1}
\maketitlesupplementary

\section{Height Initialization}
When initializing the height of the Gaussian surfel using the vehicle pose, the roll and pitch angles of the vehicle need to be taken into account and the Gaussian surfel should be in the $xy$-plane of the vehicle coordinate system. We place the vehicle coordinate system origin at the location of the vehicle's rear axle center grounding point.  

For example, the position of a Gaussian surfel in the world coordinate system is $(x_i,y_i, z_i)$. We first find the nearest vehicle position in the $xy$-plane under the world coordinate system, i.e., only the $xy$-coordinates are used in the calculation of the nearest neighbor, since the height of the Gaussian surfel is unknown at this point. If the corresponding nearest vehicle position and rotation denoted as $(x_v,y_v,z_v)$ and $\mathbf{R}_v=(\mathbf{n}_1, \mathbf{n}_2, \mathbf{n}_3)$, respectively, where $\mathbf{n}_3=(n_{31},n_{32},n_{33})^T$. The Gaussian surface is in the vehicle $xy$-plane as shown in the following equation:
\begin{equation}
\mathbf{n}_3 \cdot (x_i-x_v, y_i-y_v, z_i-z_v) = 0,
\end{equation}
Then $z_i$ can be represented as:
\begin{equation}
    z_i = z_v-\frac{1}{n_{33}}(n_{31}\cdot(x_i-x_v)+n_{32}\cdot(y_i-y_v)).
\end{equation}

\begin{figure}[t]
  \centering
  \includegraphics[width=\linewidth]{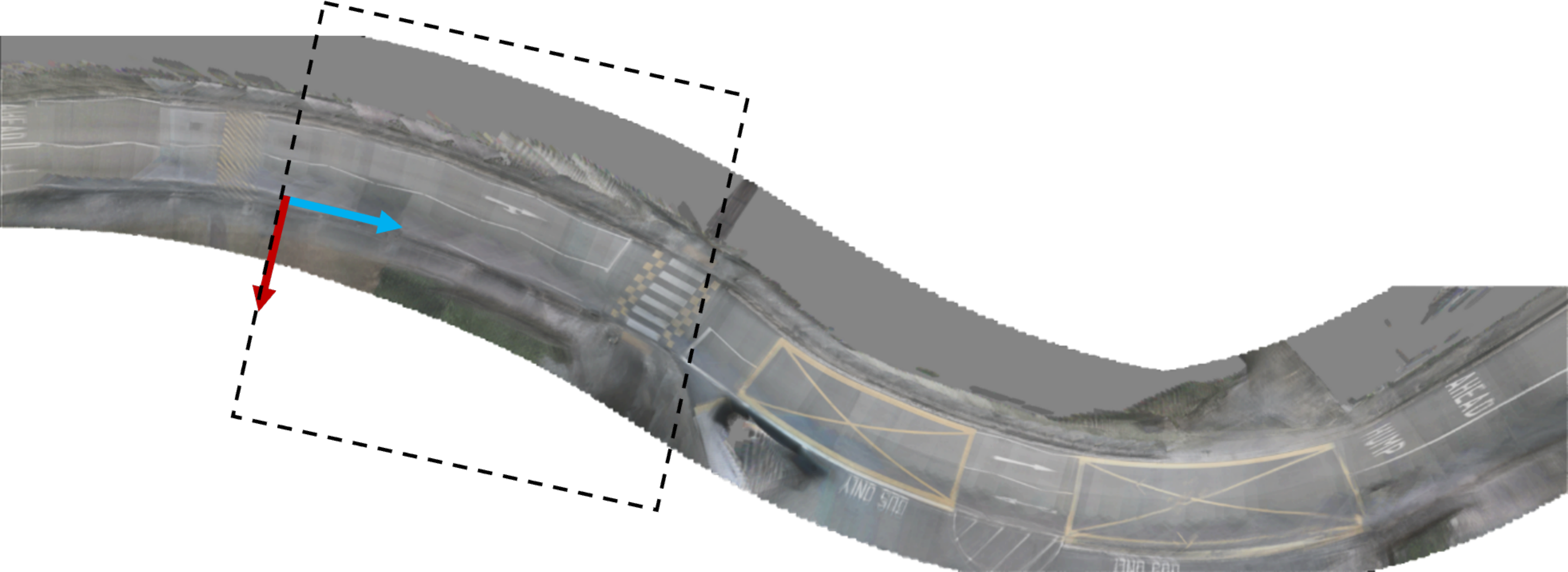}
  \caption{Perspective rendering acceleration. The blue arrows and red arrows denote the X and Z axes of the camera's coordinate system, respectively. The dashed box indicates the range of Gaussian surfels used in rendering this camera image.}
  \label{fig:per_acc}
\end{figure}

\section{Optimization Acceleration}
\label{sec:acceleration}

\subsection{Perspective Rendering Acceleration}
When rendering the colours and semantics in \cref{sec:render}, due to the large number of Gaussian surfels on the ground of a large scene, in order to speed up rendering, only Gaussian surfels within a certain range in front of the vehicle camera are used for rendering. 

As shown in \cref{fig:per_acc}, first we project the x-axis and z-axis of the camera's coordinate system onto the image plane, and then we extend it by 20m along the positive and negative directions of the x-axis, and 40m along the positive direction of the z-axis, respectively. Finally, we can obtain the rectangular box shown by the dashed line. When rendering the image of this camera, only the Gaussian surface inside the rectangular box is used.

\subsection{Orthogonal Rendering Acceleration}

When visualizing the rendered image on the BEV, this imposes a great computational burden due to the fact that we have to use all of the huge amount of Gaussian surfaces in the scene. Since the BEV image is obtained by orthogonal projection, we can render parts of the image separately by using multiple orthogonal cameras, and finally concatenate these images to get the rendered result on the BEV view. This will greatly reduce the computational burden each time.

In practice, each orthogonal camera will render an image with a resolution of 2000 pixel $\times$ 2000 pixel. The ground resolution we use is 0.05 m/pixel, i.e. each camera renders a ground range of 100m $\times$ 100m.

\begin{figure}[t]
  \centering
  \includegraphics[width=\linewidth]{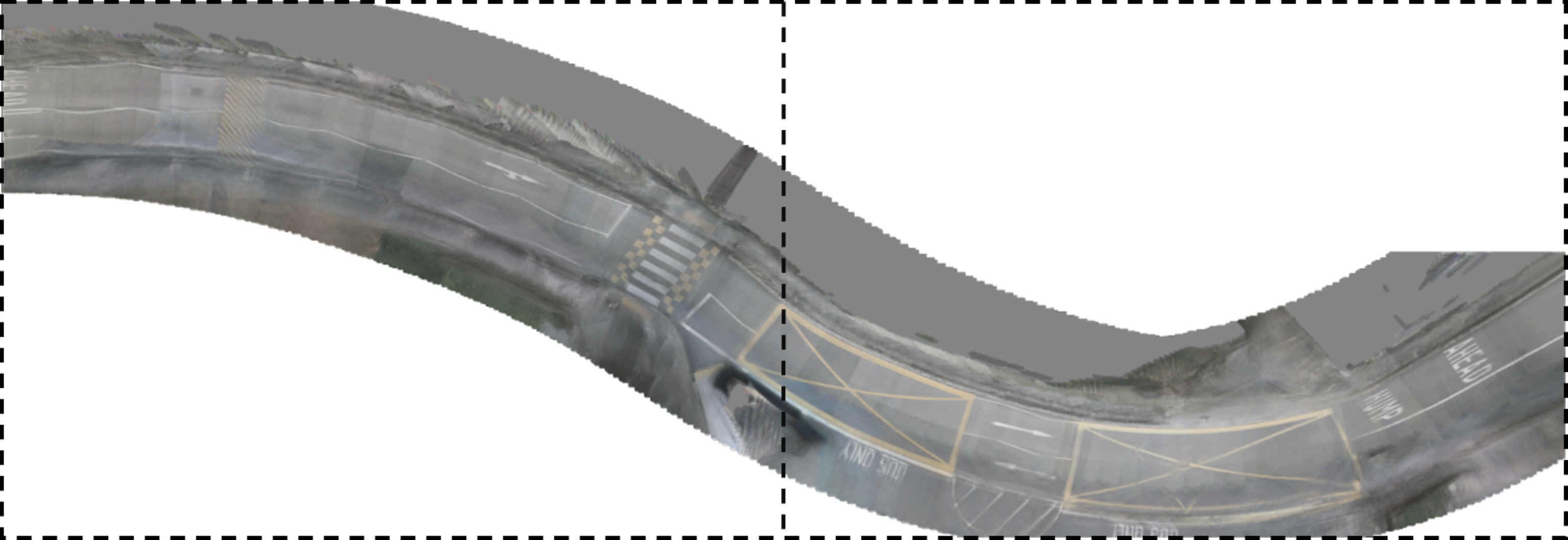}
  \caption{Orthogonal rendering acceleration. The dashed boxes indicate the orthogonal projection images rendered separately. The final BEV image is obtained by combining these separately rendered images.}
  \label{fig:bev_acc}
\end{figure}

\begin{figure*}[tbp]
  \centering
  \includegraphics[width=0.9\linewidth]{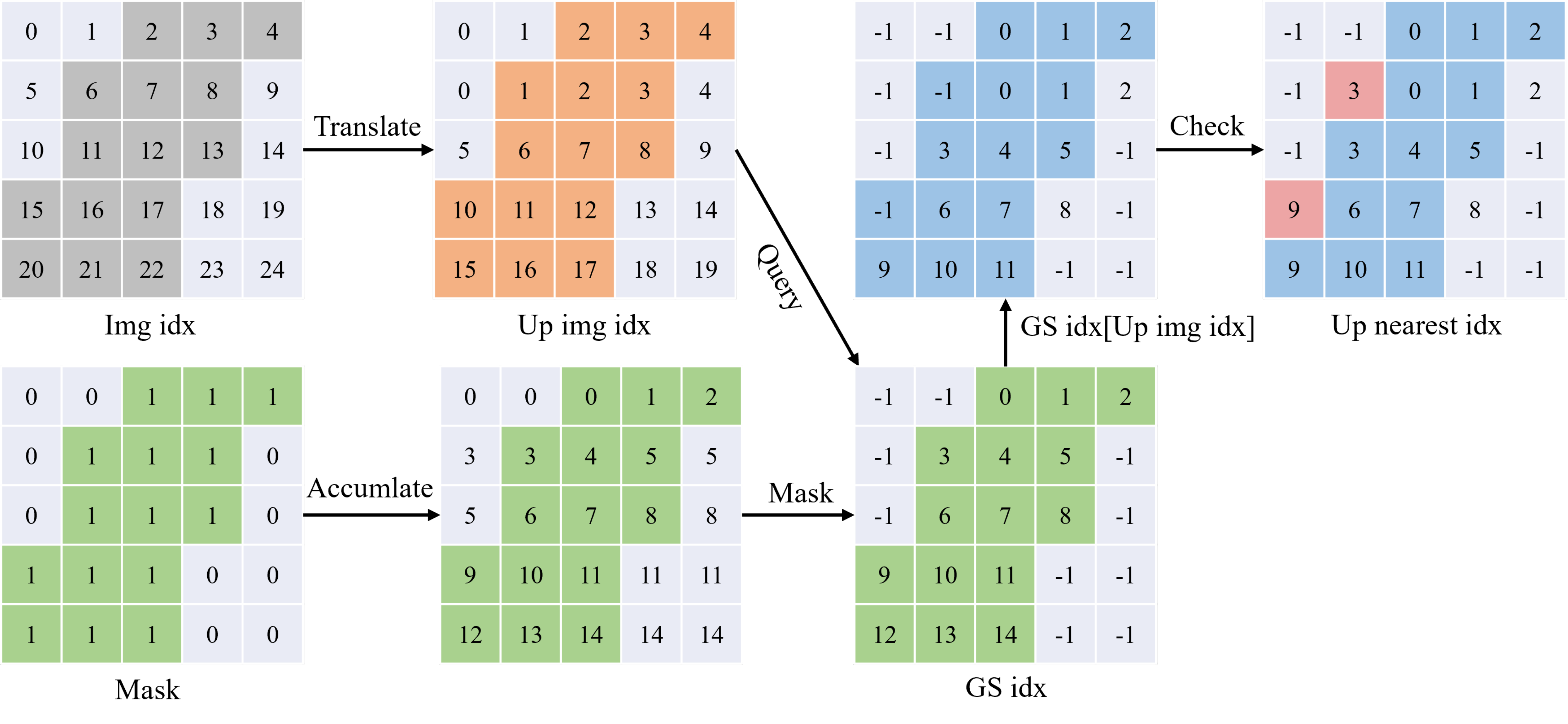}
  \caption{Smooth loss $K$ nearest neighbor acceleration.}
  \label{fig:knn}
\end{figure*}

\subsection{Smooth Loss $K$ Nearest Neighbor Acceleration}
When calculating the smooth loss in \cref{sec:loss} , it is necessary to find the K-nearest neighbours of each Gaussian surfel. Because the number of Gaussian surfels in a large scene is huge, it would consume a lot of time. To solve this case, we assign Gaussian surfels to image pixel indexes in the $xy$-plane, and compute the $K$ nearest neighbors efficiently by querying the pixel indexes. The $K$ nearest neighbour lookup process is shown in \cref{fig:knn}.

In practice, we query the nearest neighbours in the top, bottom, left and right directions. The \cref{fig:knn} shows an example of querying the nearest neighbours on the top. First, the square mesh vertices are projected onto the xy-plane, and each vertex is assigned a pixel index, as shown in the Fig. ``Img idx''.   Also, we project the position of the vehicle onto this image plane and use the image dilation algorithm to obtain the road mask, as shown in Fig. ``Mask''. We just need to place Gaussian surfels at the mesh vertices on the road surface as shown in Fig. ``GS idx''. This process can be done in a very short time using accumulation. The upper nearest neighbour of each pixel is the pixel above it in the image, as shown in Fig. ``Up img idx''. We can use ``Up img idx'' to look in ``GS idx'' to obtain the upper nearest neighbour of each Gaussian surfel. Of course, it is also necessary to check the portion of the surface that is beyond the road and correct it to its own index.

\section{Visualization}
We provide visualization results of the four additional nuScenes scenes from the experiment. Besides the 00 sequence, we also provide the visualization results for the five scenes 02,04,06,08 and 10 in the 00-10 sequences in KITTI.

\begin{figure*}[htbp]
  \centering
    \begin{subfigure}{0.48\linewidth} 
      \includegraphics[width=\linewidth]{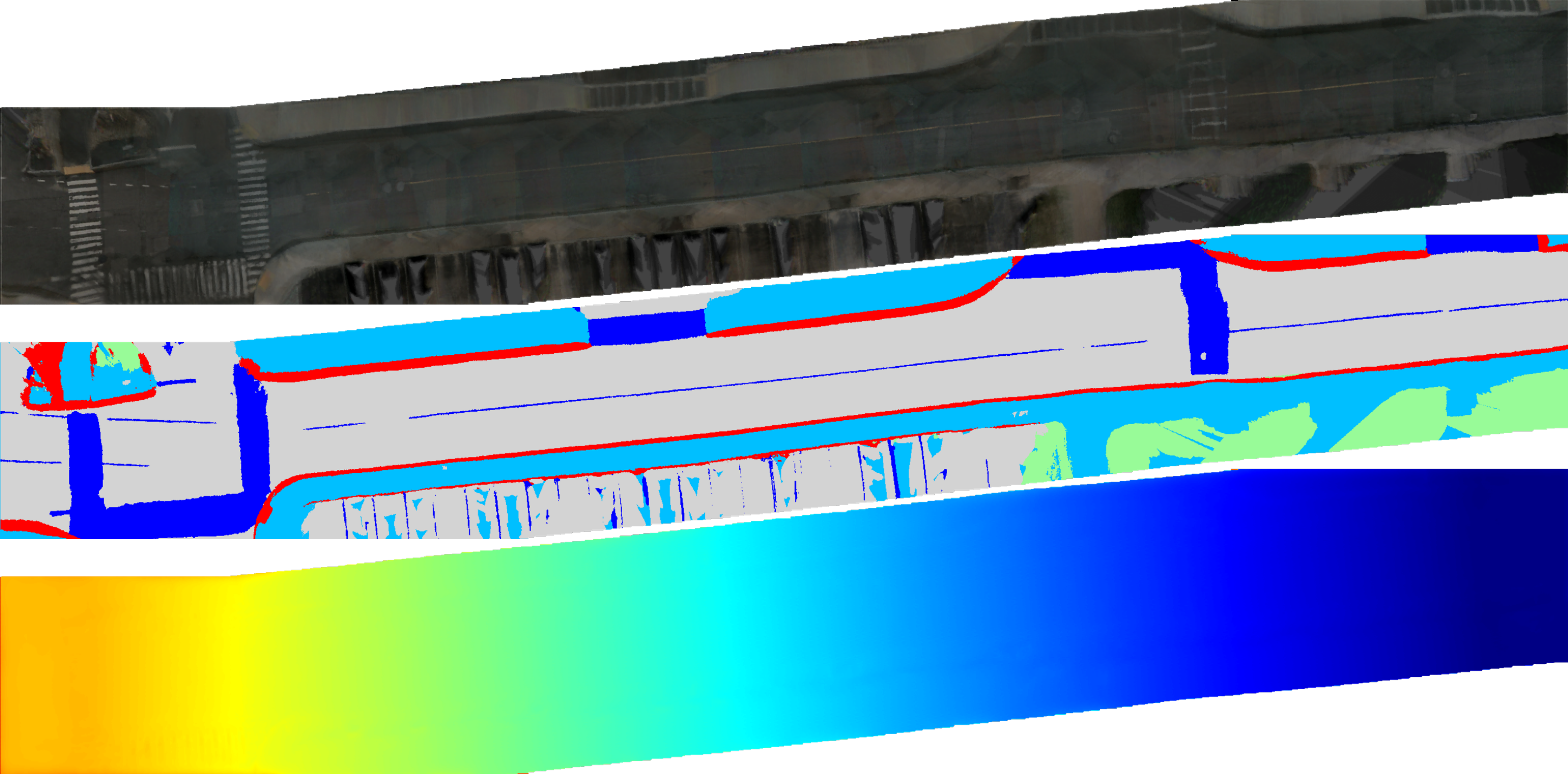}
        \caption{Without LiDAR}
        \label{fig:0064_noz}
    \end{subfigure} \begin{subfigure}{0.48\linewidth}
      \includegraphics[width=\linewidth]{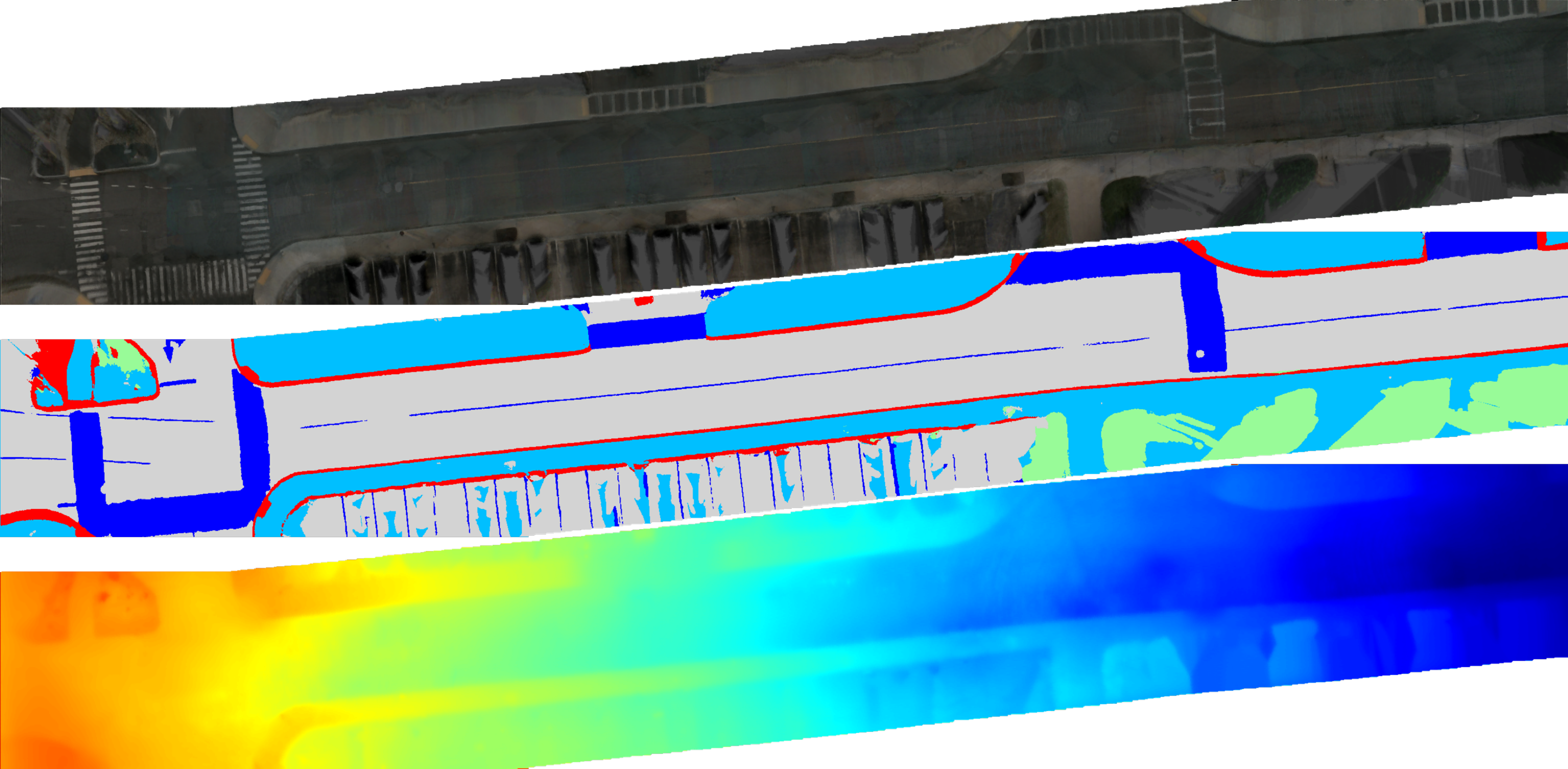}
        \caption{With LiDAR}
        \label{fig:0064_z}
    \end{subfigure}
\caption{Scene-0064}
\label{fig:0064}
\end{figure*}

\begin{figure*}[htbp]
  \centering
    \begin{subfigure}{0.48\linewidth} 
      \includegraphics[width=\linewidth]{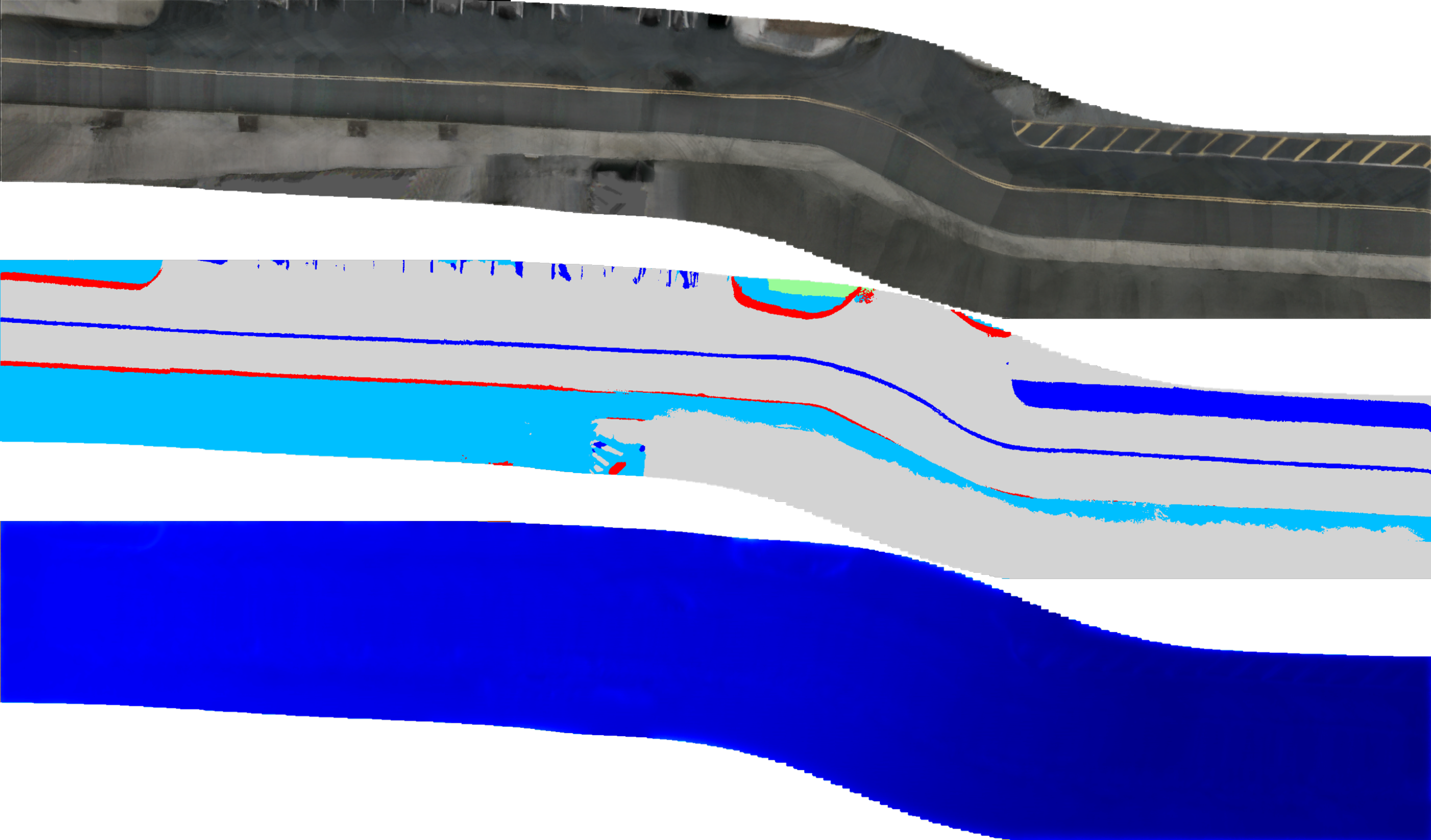}
        \caption{Without LiDAR}
        \label{fig:0212_noz}
    \end{subfigure} \begin{subfigure}{0.48\linewidth}
      \includegraphics[width=\linewidth]{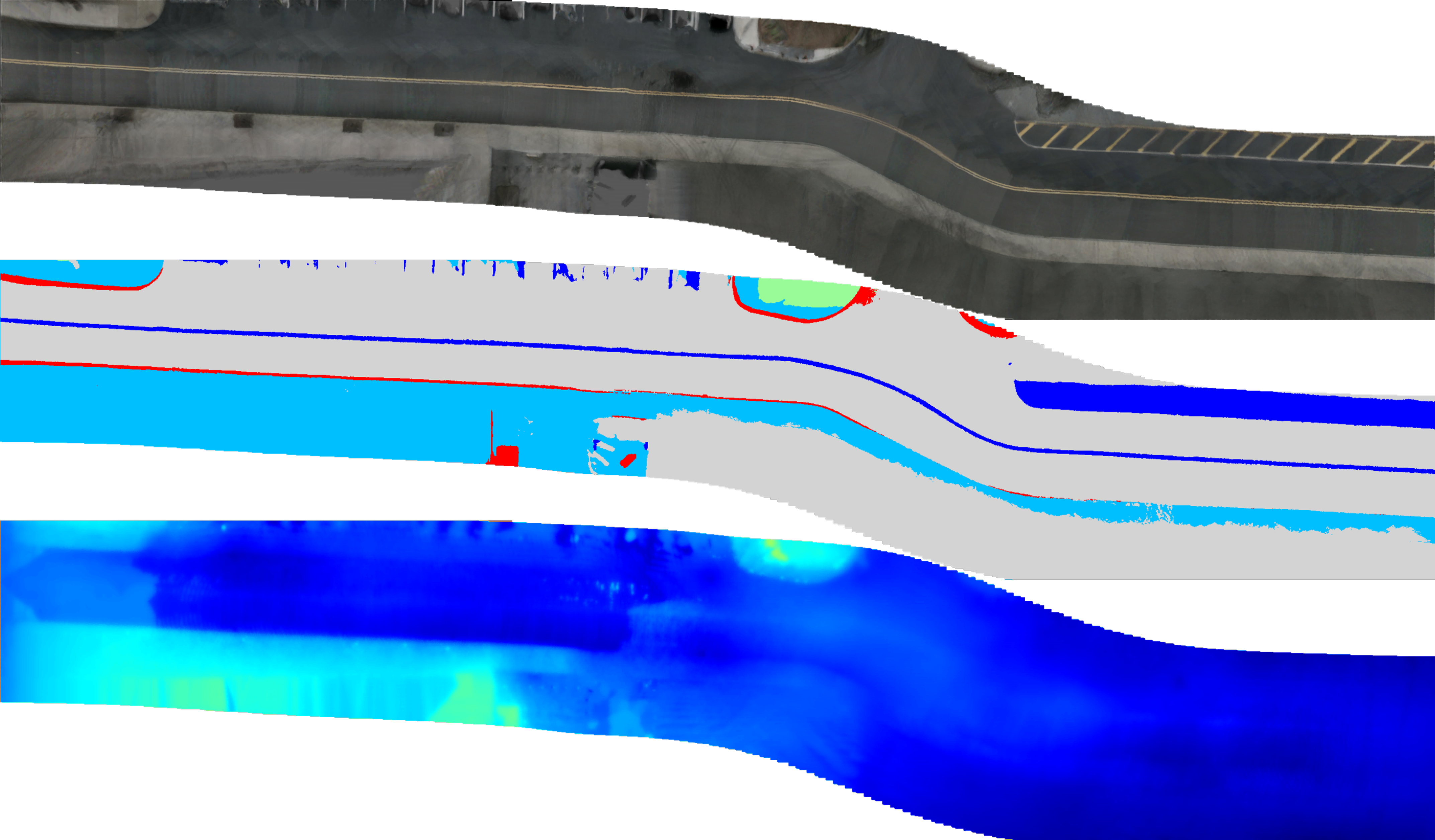}
        \caption{With LiDAR}
        \label{fig:0212_z}
    \end{subfigure}
\caption{Scene-0212}
\label{fig:0212}
\end{figure*}

\begin{figure*}[htbp]
  \centering
    \begin{subfigure}{0.48\linewidth} 
      \includegraphics[width=\linewidth]{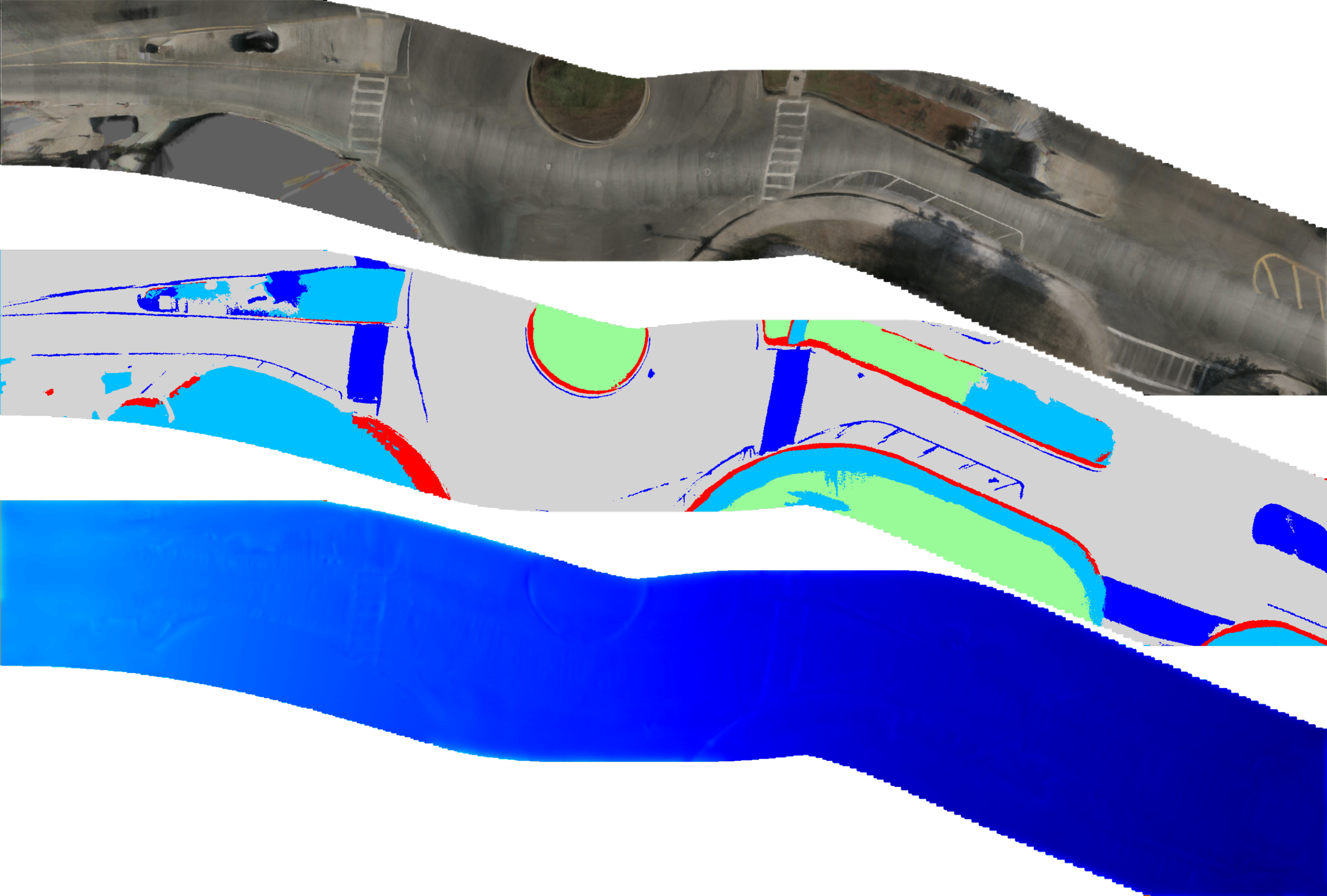}
        \caption{Without LiDAR}
        \label{fig:0523_noz}
    \end{subfigure} \begin{subfigure}{0.48\linewidth}
      \includegraphics[width=\linewidth]{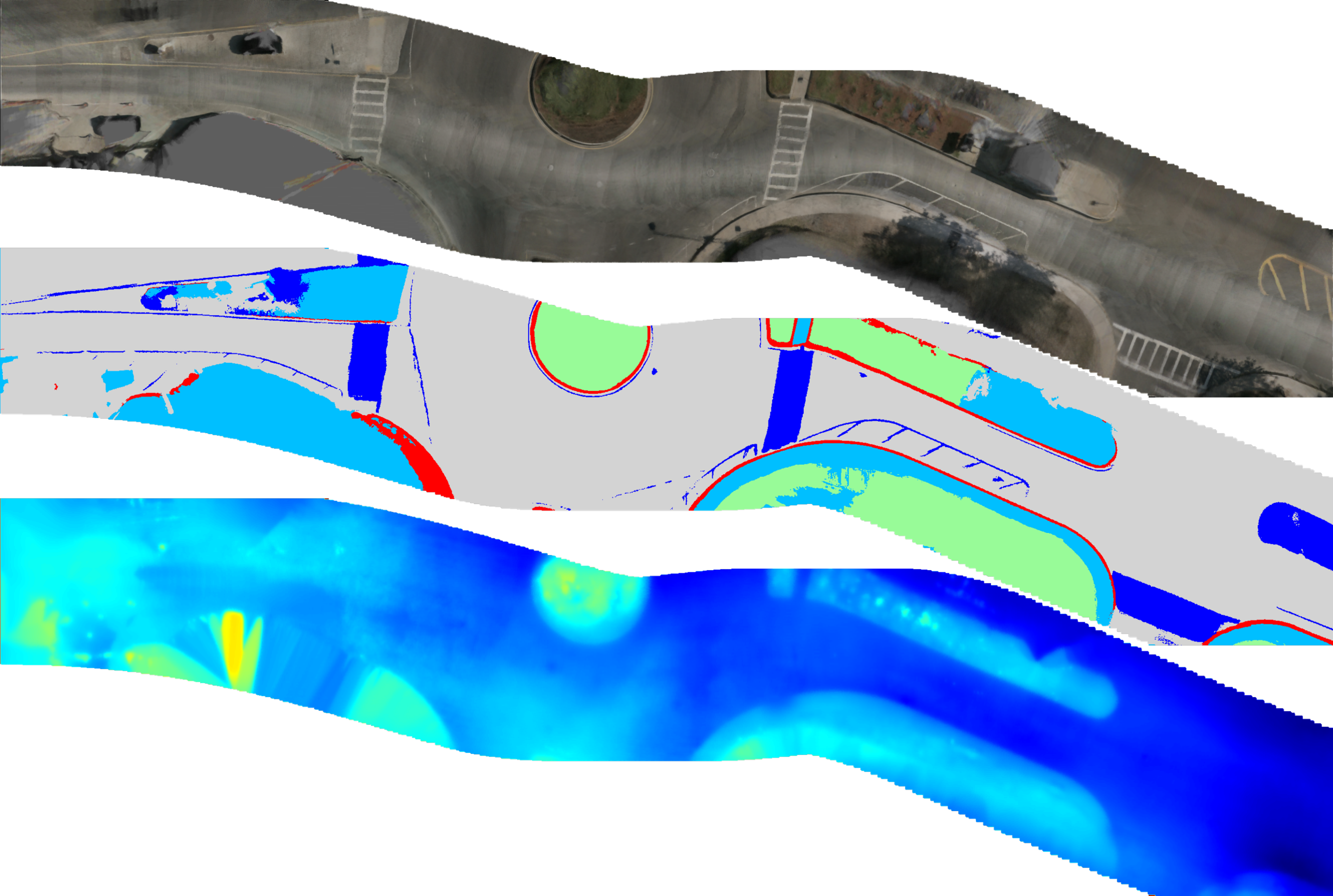}
        \caption{With LiDAR}
        \label{fig:0523_z}
    \end{subfigure}
\caption{Scene-0523}
\label{fig:0523}
\end{figure*}

\begin{figure*}[htbp]
  \centering
    \begin{subfigure}{0.48\linewidth} 
      \includegraphics[width=\linewidth]{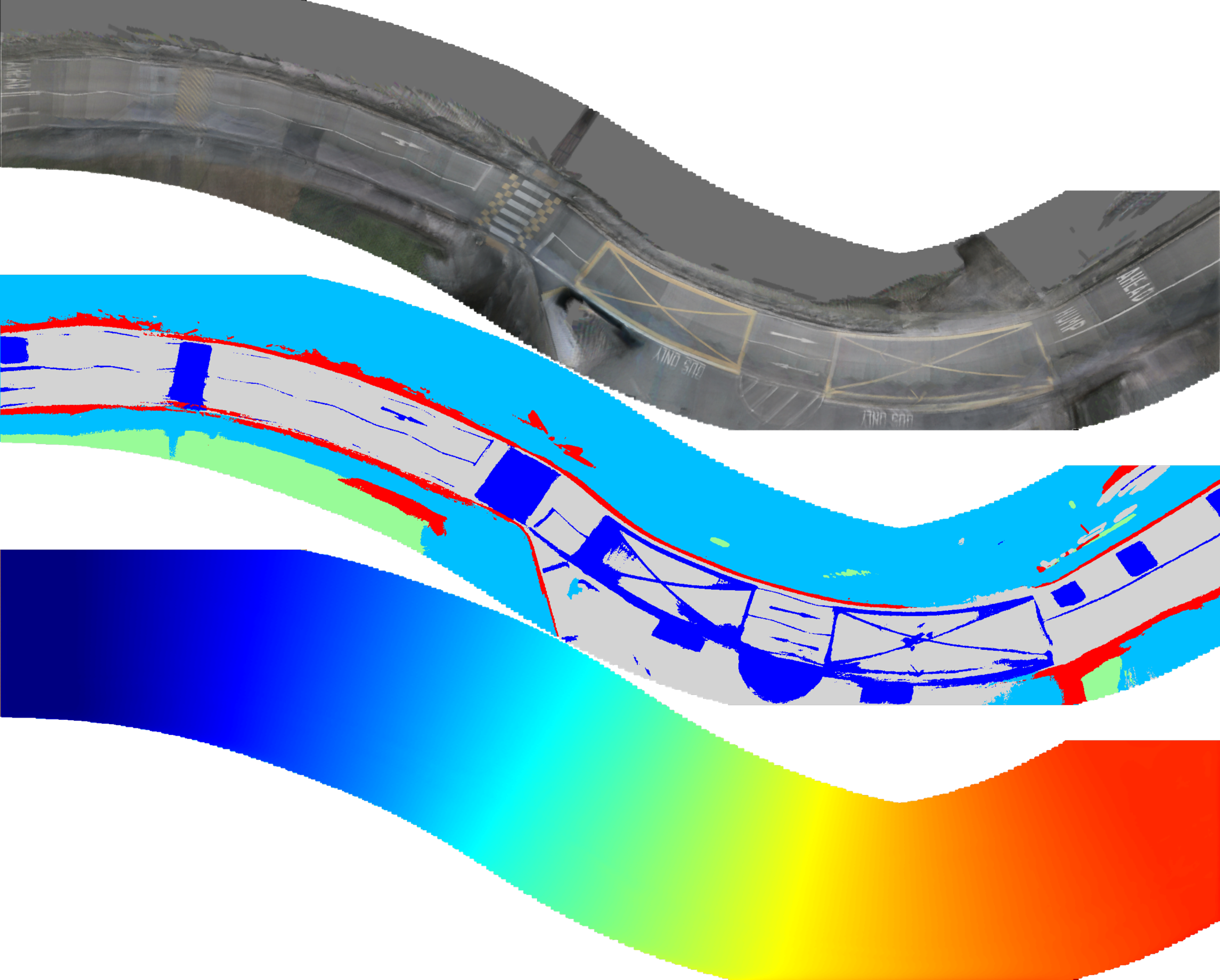}
        \caption{Without LiDAR}
        \label{fig:0856_noz}
    \end{subfigure} \begin{subfigure}{0.48\linewidth}
      \includegraphics[width=\linewidth]{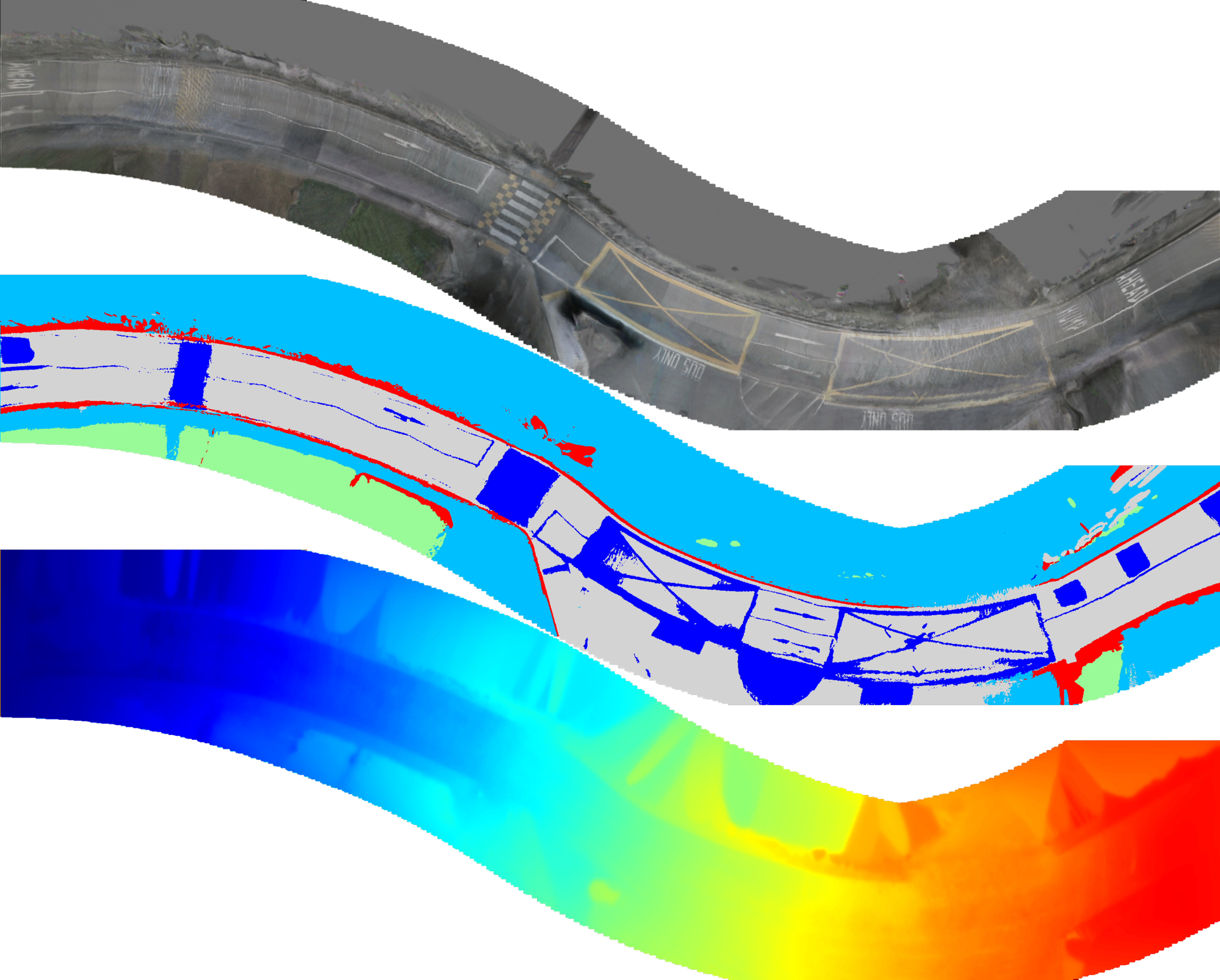}
        \caption{With LiDAR}
        \label{fig:0856_z}
    \end{subfigure}
\caption{Scene-0856}
\label{fig:0856}
\end{figure*}

\begin{figure*}[htbp]
  \centering
  \includegraphics[width=\linewidth]{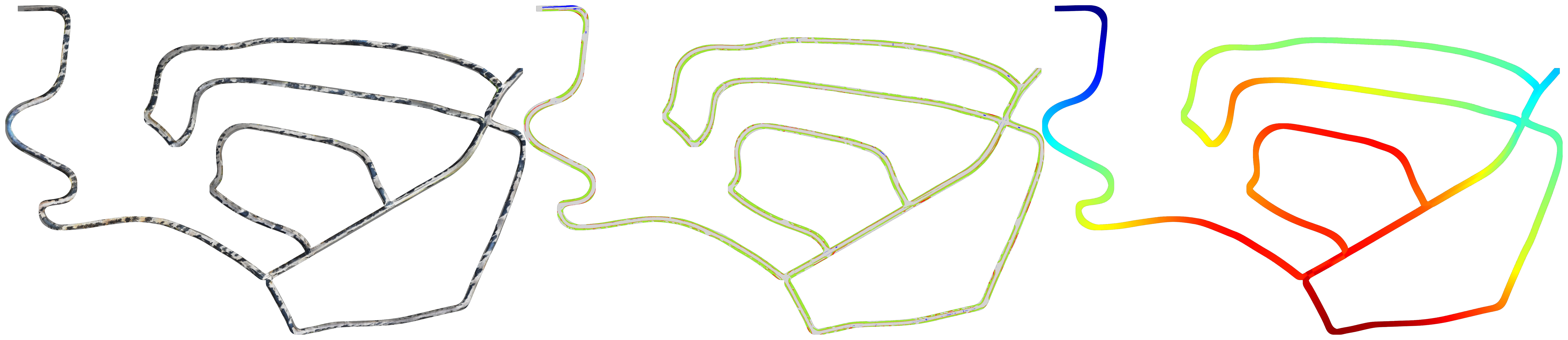}
\caption{Seq-02}
\label{fig:seq02}
\end{figure*}

\begin{figure*}[htbp]
  \centering
  \includegraphics[width=\linewidth]{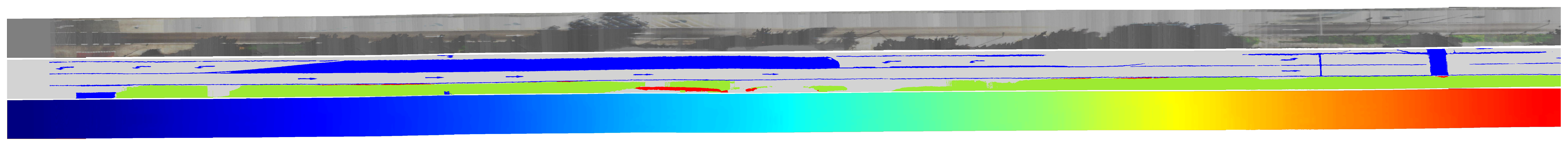}
\caption{Seq-04}
\label{fig:seq04}
\end{figure*}
\begin{figure*}[htbp]
  \centering
  \includegraphics[width=\linewidth]{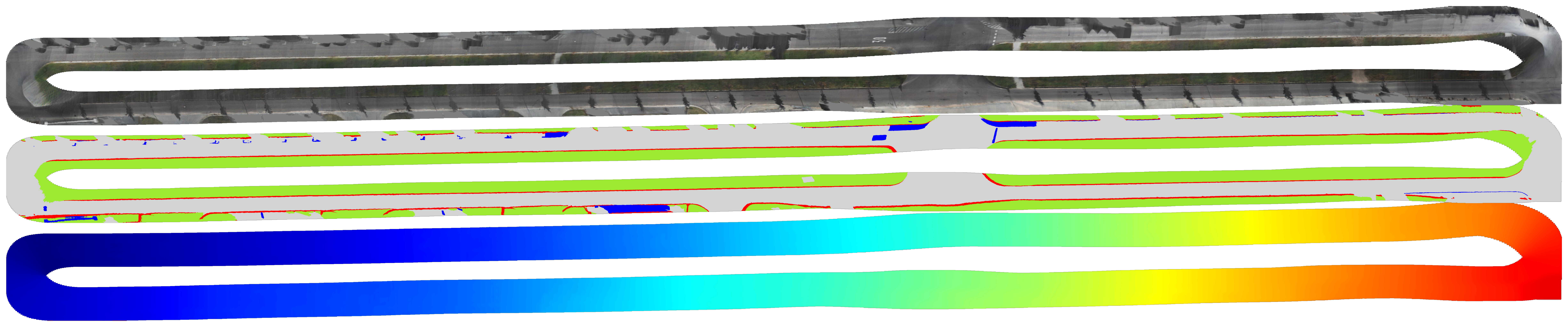}
\caption{Seq-06}
\label{fig:seq06}
\end{figure*}
\begin{figure*}[htbp]
  \centering
  \includegraphics[width=\linewidth]{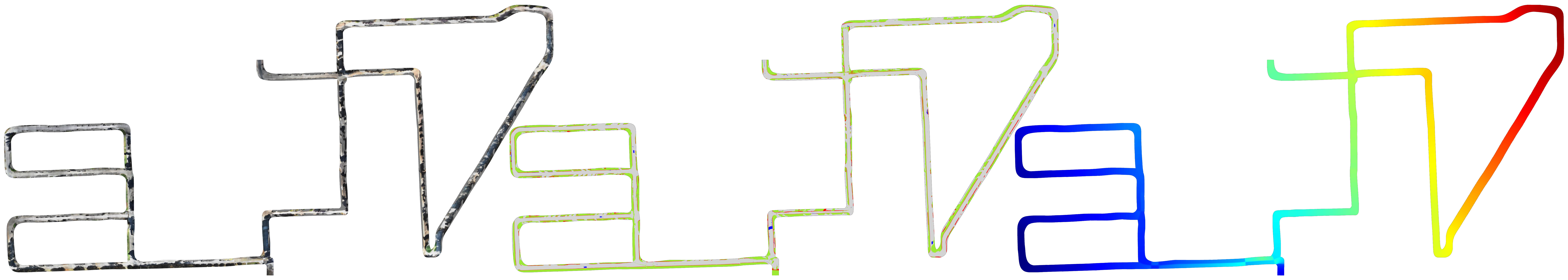}
\caption{Seq-08}
\label{fig:seq08}
\end{figure*}
\begin{figure*}[htbp]
  \centering
  \includegraphics[width=\linewidth]{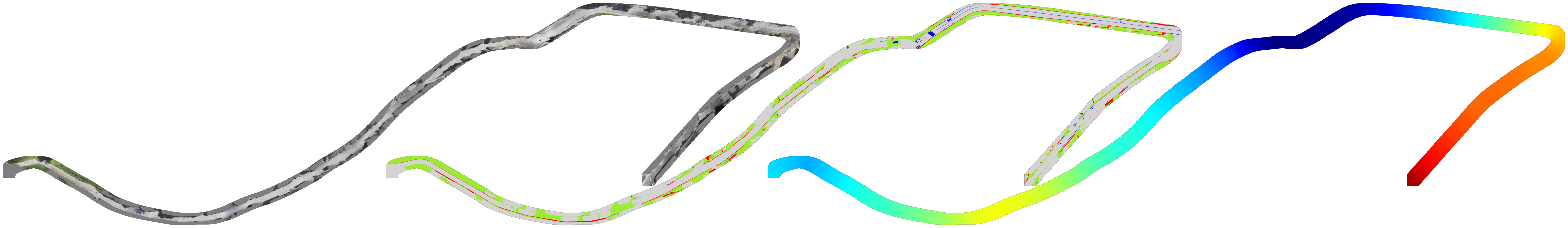}
\caption{Seq-10}
\label{fig:seq10}
\end{figure*}

% % 
% Having the supplementary compiled together with the main paper means that:
% % 
% \begin{itemize}
% \item The supplementary can back-reference sections of the main paper, for example, we can refer to \cref{sec:intro};
% \item The main paper can forward reference sub-sections within the supplementary explicitly (e.g. referring to a particular experiment); 
% \item When submitted to arXiv, the supplementary will already included at the end of the paper.
% \end{itemize}
% % 
% To split the supplementary pages from the main paper, you can use \href{https://support.apple.com/en-ca/guide/preview/prvw11793/mac#:~:text=Delete%20a%20page%20from%20a,or%20choose%20Edit%20%3E%20Delete).}{Preview (on macOS)}, \href{https://www.adobe.com/acrobat/how-to/delete-pages-from-pdf.html#:~:text=Choose%20%E2%80%9CTools%E2%80%9D%20%3E%20%E2%80%9COrganize,or%20pages%20from%20the%20file.}{Adobe Acrobat} (on all OSs), as well as \href{https://superuser.com/questions/517986/is-it-possible-to-delete-some-pages-of-a-pdf-document}{command line tools}.

\end{document}